\documentclass[preprint]{elsarticle}

\usepackage{hyperref}
\usepackage{todonotes}
\usepackage{caption}

\newcommand{\bftab}{\fontseries{b}\selectfont}

\journal{Neural Networks Journal}

\begin{document}

\begin{frontmatter}
\title{Anti-Transfer Learning for Task Invariance in Convolutional Neural Networks for Speech Processing}



\author{Eric Guizzo\footnote{Corresponding author.}, Tillman Weyde, Giacomo Tarroni}
\address{Department of Computer Science \\ 
City, University of London \\ 
\texttt{\{eric.guizzo,t.e.weyde,giacomo.tarroni\}@city.ac.uk}}




\begin{abstract}

We introduce the novel concept of \emph{anti-transfer} learning for speech processing with convolutional neural networks. 
While 
transfer learning assumes 
that the learning process for a target task will benefit from re-using representations learned for another task, anti-transfer  avoids the learning of representations that have been learned for an \emph{orthogonal} task, i.e., one that is not relevant and potentially misleading for the target task, such as speaker identity for speech recognition or speech content for emotion recognition. 

In anti-transfer learning, we penalize similarity between activations of a network being trained and another one previously trained on an orthogonal task, which yields more suitable representations. 
This leads to better generalization and provides a degree of control over correlations that are spurious or undesirable, e.g. to avoid social bias. 
We have implemented anti-transfer for convolutional neural networks in different configurations with several similarity metrics and aggregation functions, which we evaluate and analyze with several speech and audio tasks and settings, using six datasets. 
We show that anti-transfer actually leads to the intended 
invariance to the orthogonal task and to more appropriate features for the target task at hand.
Anti-transfer learning consistently improves classification accuracy in all test cases. 

While anti-transfer creates computation and memory cost at training time, there is relatively little computation cost when using pre-trained models for orthogonal tasks.
Anti-transfer is widely applicable and particularly useful where a specific invariance is desirable
or where trained models are available and labeled data for orthogonal tasks are difficult to obtain. 

\end{abstract}


\end{frontmatter}


\section{Introduction}
\label{intro}
In recent years, transfer learning has become a popular method in speech and audio processing to make use of existing deep learning models that have been trained on large datasets. 
The assumption underlying transfer learning is that the internal representations learned to solve one task will be relevant for another task. 
This can improve the performance of a model in terms of training time and overall accuracy even across tasks and domains, 
and has been proven to be particularly useful in cases when data availability for the target task is limited \cite{DBLP:conf/ismir/OordDS14,DBLP:conf/naacl/BansalKLLG19, DBLP:conf/ismir/DielemanBS11, DBLP:journals/corr/abs-1909-10924}.

We introduce here the concept of anti-transfer learning, which is based on the idea that if a neural network can be used to teach another network what to do, it may also be used to teach what \textit{not} to do. 
Based on the observation that some tasks may be irrelevant and confounding or undesirable to influence the target task, we try to avoid representations learned for one task when learning to solve another.  
We call the task which should not influence the predictions an \textit{orthogonal} task, as our intention is that the predictions of our target should be independent of it. 
What constitutes an orthogonal task depends on the nature of the tasks and the intention of the user. 
We see two main application scenarios: first, improving generalization by discouraging reliance on spurious associations, e.g., word recognition and speaker identity, and second, discouraging undesirable bias, e.g. that gender or ethnicity should not influence financial decisions. 

In this paper we focus on the first scenario, and particularly on audio applications. 
Spurious correlations occur frequently in real-world data and are sometimes unavoidable. E.g., we expect the word `joy' to be associated with a happy expression in natural speech.
This association may be useful to resolve ambiguities, but a model overly reliant on this  may not generalize in cases where the association does not hold, e.g. the word 'joy' pronounced with a sad expression. 
Similarly, the frequency of word use is not equally distributed between different speakers, genders or ethnicities, but we would prefer our models not to depend on these features when they recognize words, both in the interest of generalization and in avoidance of bias or stereotyping. 
This problem could be addressed by creating or collecting more data, that contains all variants of emotional expressions for all words, or all words uttered by all speakers but this not practical in general. 
However, with anti-trarnsfer we can discourage the use of emotional features for word recognition, or speaker identity for emotion recognition, respectively, and thus avoid that dependency and improve generalization from limited datasets.  





Anti-transfer can be used to address open research problems in speech and audio processing, such as speaker or context invariance in word or emotion recognition \cite{DBLP:conf/interspeech/KitzaSN18,DBLP:conf/interspeech/MildeB18,DBLP:conf/icira/LiuXLH19,DBLP:conf/asru/JalalMH19,DBLP:journals/amcs/RybkaJ13,speakerdependent2013}. 
In our experiments, we compare anti-transfer learning to regular transfer learning and learning from scratch on speech and music audio tasks. 
A common approach for transfer learning with deep learning models is to use a pre-trained network as starting point through weight initialization, i.e. re-training a pre-trained network or part of it \cite{DBLP:conf/icann/TanSKZYL18}. 
Support for this approach is built into popular machine learning libraries, such as \textit{Tensorflow}\footnote{\url{https://www.tensorflow.org/}} and \textit{PyTorch}\footnote{\url{https://pytorch.org/}}, along with models pre-trained on disparate tasks.
In anti-transfer learning, we penalize instead  the use of features that have been learned for the orthogonal task when training for the target task. 
Our results show that this leads to greater invariance to the target predictions from the orthogonal task
and improves the generalization of the models.
 
The specific contributions of this work are the following:
\begin{itemize}
 \item For the first time, to the best of our knowledge, we introduce the concept of \emph{anti-transfer} learning to achieve task-invariance between a pre-trained network and a new one.
 \item We implement \emph{anti-transfer} learning for convolutional neural networks (CNNs) with a number of different similarity measures and aggregation functions. 
 The source code is publicly available\footnote{\url{https://github.com/ericguizzo/anti_transfer}}.
 \item We demonstrate the effectiveness of \textit{anti-transfer} learning 
 for speech and audio by evaluating it on two speech-related tasks and one music-audio task, using six different datasets in several configurations. 
 We achieve improvements in all tasks over non-transfer and standard transfer learning.
 
 \item We provide results of ablation studies and visualizations to analyze the properties of anti-transfer learning. 
\end{itemize}

The remainder of this paper is organized as follows: Section~\ref{previouswork} contains a review of relevant background literature, Section~\ref{method} introduces the concept and implementation of anti-transfer learning, Section~\ref{sec:experimental} presents a performance evaluation of anti-transfer, followed by further analysis and discussion in Section~\ref{sec:analysis} 
and Section~\ref{conclusions} draws the conclusions from this paper.

\section{Related Work}
\label{previouswork}

Transfer learning has been used with neural networks for a long time and in many different applications \cite{caruana1995learning, bengio2012deep,41530, shin2016deep, tan2018survey}.
Pre-training models has become standard practice in image classification and related tasks \cite{DBLP:conf/icdar/StuderAPGK0LI19, DBLP:conf/eccv/XieR18, DBLP:journals/amm/HanJMX18}
and pre-trained language models have become a common starting point in NLP \cite{DBLP:journals/corr/abs-2003-08271}.
The transfer of knowledge from a trained network to a new task by re-using weights of a layer has been developed early on   \cite{Pratt-1992-Discriminability-Based,Gutstein-et-al-2007-Knowledge-Transfer}. 

\paragraph{Selective representation transfer} 
The approach presented in this paper is inspired by the work of \cite{gatys2016image} on style transfer on images and re-uses elements of that work. 
Based on the assumption that features 
become increasingly task-specific towards 
the last layer of a network\cite{yosinski2014transferable}, a strategy was developed by \cite{gatys2016image} to separate content and style of an image and to transfer the style alone to another image. 
The authors used a CNN that was pre-trained on object recognition as a feature extractor to estimate the style-related and the content-related information of an image in a CNN. 
The style of an image is represented by the Gram matrix computed on the initial layers, which contains information about texture, i.e. the co-occurrence of low-level features.
The content is represented by the raw feature maps of the final layers. 
During the training of the style transfer network, the feature extractor separately extracts the style and the content from two different images and compares them 
to the corresponding features extracted from an image that is being generated, creating two deep feature loss values: style and content loss. 
The minimization of these losses promotes the generation of an image with the style of one image and the content of the other one.
This idea received much attention in the computer vision community and has been further explored and improved \cite{ulyanov2016texture, dumoulin2016learned}.
It has also been applied in %
the audio domain to audio style transfer with MelGan \cite{pasini2019melgan}, using both speech and music sources. 

\paragraph{Deep feature losses} 
Deep feature losses have been used in several computer vision tasks as texture synthesis \cite{gatys2015texture}, image super-resolution \cite{johnson2016perceptual} and conditional image synthesis \cite{chen2017photographic,dosovitskiy2016generating}.
According to recent studies \cite{zhang2018unreasonable,doersch2015unsupervised}, deep feature losses are highly correlated to human perceptual judgements and are well suited to solve tasks related to semantic properties of data.
Deep feature losses have several successful applications also in the audio domain. 
They have been used by \cite{beckmann2019speech} to enhance the similarity between the deep representations of two networks and therefore transferring knowledge from one to the other, enhancing the networks' performance in several speech processing tasks. 
A deep feature loss was successfully used by \cite{sahai2019spectrogram} to perform audio source separation, obtaining a superior performance compared to spectrogram-based loss.
\cite{kegler2019deep} applied the same conceptual idea to speech enhancement, language identification, speech, noise and music classification, and speaker identification.

The majority of studies regarding deep feature losses are based on the idea of encouraging a network to develop similar deep representations of a pre-trained network in selected layers, e.g. in line with the work of \cite{gatys2016image}.
However, our main idea to perform the opposite. That is: discouraging specific deep representations that have been particularly useful for a task that is irrelevant for a target task and should thus be avoided when training for the target task, in order to not develop spurious correlations.
We actually address similar problems with a similar approach as \cite{beckmann2019speech} and \cite{kegler2019deep} but 
while they maximize the similarity 
with 
representation of a pre-trained network, our aim is to minimize it.
There are also substantial differences in the implementation as we use Gram aggregation and a different similarity measure, as explained in Section \ref{method}.

\paragraph{Feature diversity}
Minimizing feature similarity has been shown earlier to improve robustness and generalization. 
In the context of ensemble models, \cite{yao2004evolving} minimized mutual information between neural networks. 
More recently, the minimum hyperspherical energy (MHE) regularization was introduced by \cite{liu2018learning} and applied to audio source separation by \cite{perez2020improving}. MHE encourages diverse weight vectors within a network to improve generalization, but it differs from our approach since we encourage dissimilarity of feature maps and with respect to another model. 

\paragraph{Domain adaptation} 
A common use case for transfer learning is domain adaptation, e.g. to different recording equipment or environments, and a common approach is to maximize the feature invariance to the domain of the data. 
\textit{Mutual Information Minimization} is used in \cite{wang2019learning} to extract features independent from the domain of the data points by maximizing the feature invariance to their domain indicator.  
This is different from our approach in terms of applications, as we are training to transfer between tasks, within or between domains. 
However, when viewing the domains as orthogonal tasks, we can compare domain adaptation to anti-transfer. 
In Domain Adversarial Training (DAT) \cite{DBLP:journals/jmlr/GaninUAGLLML16}, a gradient reversal layer is introduced to maximize the loss on domain identification while minimizing the classification loss. A similar approach, but with a  Siamese architecture, is introduced in \cite{DBLP:conf/iccv/MotiianPAD17}. 
In \cite{DBLP:conf/iclr/TzengHSD17}, a more general framework is presented, including generative adversarial approaches, that is also applied in domain adaptation for acoustic scene classification using unlabeled data for the target domain \cite{DBLP:conf/waspaa/DrossosMV19}. 
There are two notable differences between these approaches and ours: first, we 
directly compare the feature activations in our loss function as opposed to propagating gradients derived from domain labels,  
and second, most of these approaches require labeled data from the source domain (analog to our orthogonal task), while anti-transfer only requires a pre-trained model, which does not have to be trained on the same dataset. 

\paragraph{Disentanglement} 
The representation of independent properties of objects or processes has been recently explored in the literature and usually referred to as \emph{disentanglement} \cite{DBLP:journals/corr/abs-1811-03271, DBLP:conf/interspeech/ChouYLL18,DBLP:conf/icassp/NagraniCAZ20,DBLP:conf/icassp/LeeBSJN20}.  
Methods for achieving disentanglement include  adversarial training  \cite{DBLP:journals/corr/abs-1904-04772}
or specific architectures, such as partitioned or factorized variational autoencoders \cite{DBLP:conf/icml/LiM18a,DBLP:journals/corr/abs-1805-11264}.
Anti-transfer can be considered a special case of disentanglement, aiming at the invariance to the internal representations of distinct orthogonal models.

\section{Method}
\label{method}

The main idea of anti-transfer learning is to encourage dissimilarity of a model's deep representations 
with respect to another 
model 
with the same architecture but pre-trained on an orthogonal task. 
We focus here on CNNs which have been immensely popular in recent years and achieve state of the art results on many audio tasks, e.g. \cite{DBLP:conf/ismir/JanssonHMBKW17, DBLP:conf/interspeech/NeumannV17, DBLP:journals/spl/SalamonB17}. 

\subsection{Approach}
We achieve anti-transfer learning through the introduction of 
an anti-transfer loss term during training, that is a deep feature loss \cite{dosovitskiy2016generating}.
The anti-transfer loss
measures the 
similarity between the deep representations that the network is learning and a 
pre-trained network 
with the same architecture.
By adding this term as a penalty to the loss function we encourage the trained network to develop deep representations that are different from the pre-trained network.
In other words, we encourage the  network being trained to develop feature representations that are good for its target task but different from those developed to solve the orthogonal task in the pre-trained network.
This reduces the trained network's dependency on the 
orthogonal task's classes, e.g. the dependency of word recognition on speaker identity. 


\begin{figure}[tb]
  \centering
  \includegraphics[width=\textwidth]{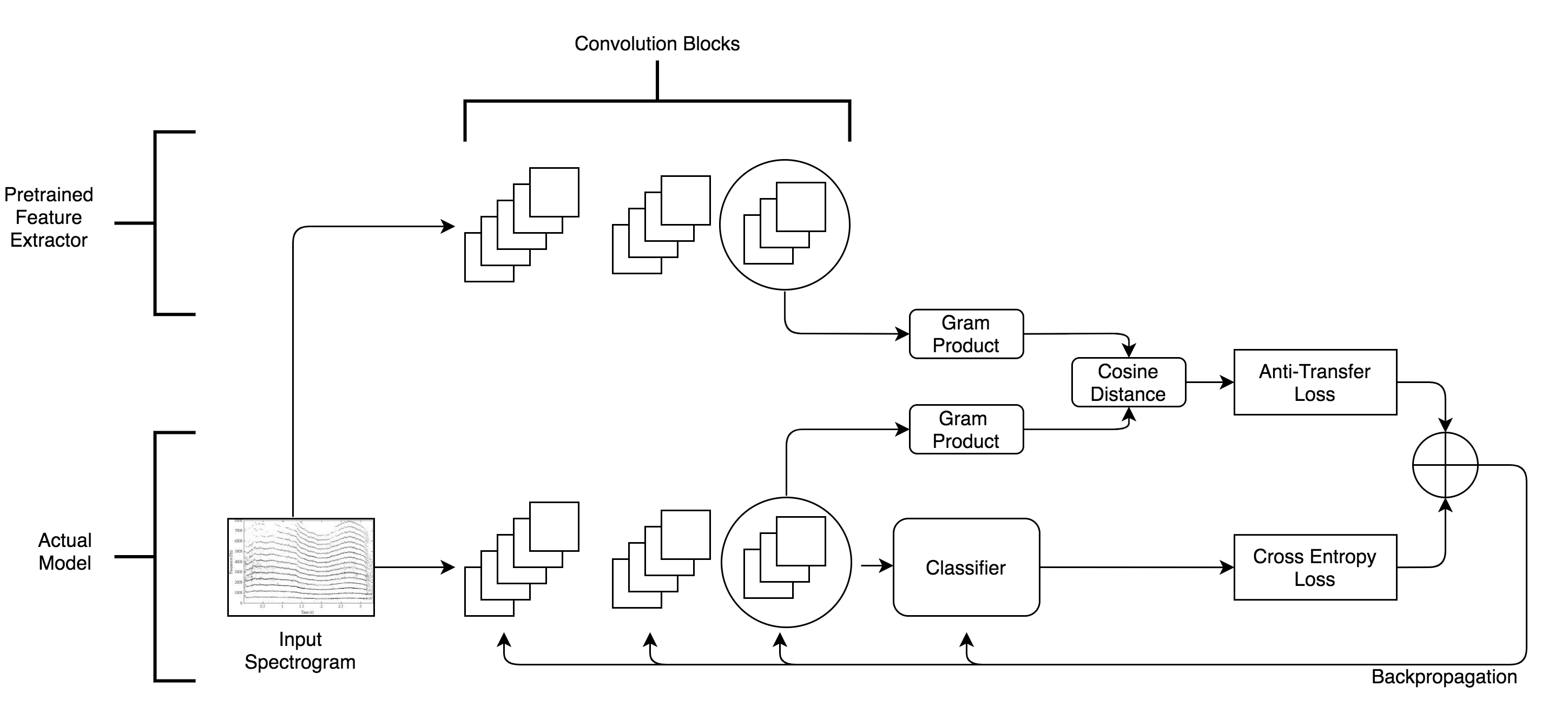}
\caption{Block diagram of a CNN network with \textbf{anti-transfer learning} applied to a classification task. We use spectrograms of audio signals as the input, but anti-transfer is not specific to the audio domain or spectrogram representations.}
\label{fig:diagram}
\end{figure}

Figure \ref{fig:diagram} depicts a block diagram of a generic CNN with  \textit{anti-transfer} learning applied. 
As the diagram shows, this architecture has two parallel networks: a \textit{pre-trained feature extractor} (in the upper part), which is the convolutional part of the pre-trained network, with non-trainable weights and the CNN classifier that is currently being trained (in the lower part). 

\begin{table*}[tb]
\caption{The \textbf{VGG16} architecture. In this example configuration the network has an input dimension of 244x244x1 and 1000 output classes.
    }
  \centering
  \vspace{.5mm}
  \begin{tabular}{l|c|c|c|c|c}
    \bftab {Layer} & \bftab {Channels}  & \bftab {Size} & \bftab {Kernel}&\bftab {Stride}  & \bftab {Activation} \\ \hline \hline 
    Input & 1  & 244x244 & - & - & - \\
    2x Convolution & 64  &224x224 &3x3 &1 &relu \\
    Max Pooling & 64  &128x128 &3x3 & 2 &relu \\
    2x Convolution & 128  &224x224 &3x3 &1 &relu \\
    Max Pooling & 128  &56x56 &3x3 & 2 &relu \\
    2x Convolution & 256  &56x56 &3x3 &1 &relu \\
    Max Pooling & 256  &28x28 &3x3 & 2 &relu \\
    3x Convolution & 512  &28x28 &3x3 &1 &relu \\
    Max Pooling & 512  &14x14 &3x3 & 2 &relu \\
    3x Convolution & 512  &14x14 &3x3 &1 &relu \\
    Max Pooling & 512  &7x7 &3x3 & 2 &relu \\
    Fully Connected &- &25088 &- &- &relu \\
    Fully Connected &- &4096 &- &- &relu \\
    Fully Connected &- &4096 &- &- &relu \\
    Output &- &1000 &- &- &softmax \\
    
    \hline
    \hline
  \end{tabular}
  \label{table:vgg}
\end{table*}

Our implementation is based on the VGG16 Architecture \cite{simonyan2014very}, a deep CNN, with details shown in Table \ref{table:vgg}.
We selected this architecture since it has been proven to be effective in computing a deep feature loss in the audio domain \cite{beckmann2019speech}.
Nevertheless, the same concept and implementation can be translated to any other CNN design. 

\subsection{Anti-Transfer Loss}

The anti-transfer loss is computed in the forward pass. 
The input data, a spectrogram in our experiments, is forward propagated in parallel through both networks. 
The feature maps of the $n^{th}$ convolution layer in both networks are extracted and aggregated in the channel-wise Gram matrix $G$, which is computed for each network, similarly to the approach used by \cite{gatys2016image} to compute the style matrix of an image. 
The Gram matrix is computed as the inner product between the vectorized feature maps $F$ for each pair of channels:
\begin{equation}
   G_{ij} = F_i \cdot F_j . 
   \label{eq:gram}
\end{equation}
where $i,j$ are the channel numbers. 
The Gram matrix correlates the information of each channel pair over all points $x,y$,  consequently reducing the dimensionality of a feature map from 3 dimensions, ($c$, $x$, $y$), 
to 2,
($c$, $c$), where 
$c,x,y$ are the number of channels, rows and columns, respectively. 
We then calculate the anti-transfer (AT) loss $L_{AT}$ as a scalar coefficient $\beta$ multiplied by the squared cosine similarity of the vectorized Gram matrices $G_p$ (for the pre-trained net) and $G_t$ (for the net being trained):
\begin{equation}
    L_{AT} =  \beta \left(\frac{G_p \cdot G_t}{||G_p|| \, ||G_t|| }\right)^2.
    \label{eq:cos}
\end{equation}
The aggregation with the Gram matrix serves to compare all possible channel combinations at once, using a limited amount of memory.
This is essential for consistently measuring the similarity of the feature maps, where permutations can occur along the channel dimension. 
We choose the squared cosine similarity since it is naturally limited in the  interval [0,1] and therefore it can have only a limited impact in the overall loss function.
Moreover, we square it to apply a stronger penalty when the similarity is high and we re-scale by the coefficient $\beta$ as an hyperparameter to fine-tune the performance of AT learning.

The diagram in Figure \ref{fig:diagram} shows the AT loss calculated on the last convolution layer, but it is possible to apply the the AT loss to any of the convolution layers. 
Furthermore, it is possible to combine the AT loss of multiple layers in the same training, summing their AT loss values.
The total AT loss is added to the standard loss function during the training of the network
(cross entropy in our case, but AT can be used with any loss function).

The complete objective function we minimize per datapoint is therefore:
\begin{equation}
    L_{TOT} = - \sum_{i=1}^{n} t_i \log (p_i) + 
    \sum_{s \in S_{AT}} 
     L_{AT s}
\end{equation}
where $n$ is the number of classes, $t_i$ is $1$ if $i$ is the true class and $0$ otherwise, $p_x$ is the predicted probability of class $i$, $S_{AT}$ is the set of convolution layers where anti-transfer is computed, and $L_{AT s}$ is the anti-transfer loss computed for convolution layer $s$.

\subsection{Variations}\label{sub:advanced_settings}

As we present in Section \ref{sec:analysis}, we test several aggregation strategies and similarity measures.
The best combination is Gram matrix aggregation and squared cosine similarity, which is detailed above.
Different aggregation and similarity functions can be used by adapting equations~\ref{eq:gram} and~\ref{eq:cos}.

Moreover, we combine two orthogonal tasks in \textit{dual AT} loss.
To achieve this, we first train a model with anti-transfer for one orthogonal task. 
We use the result of that training to initialize the weights of a new model, which is then trained with anti-transfer on the second  orthogonal task.
It is worth noting that we apply the weight initialization to all convolution layers at once, while we apply anti-transfer to only one convolution layer per experiment. 

\section{Experimental Set-up and Results}
\label{sec:experimental}

\begin{figure}[!tb]
  \centering
  \includegraphics[width=\textwidth]{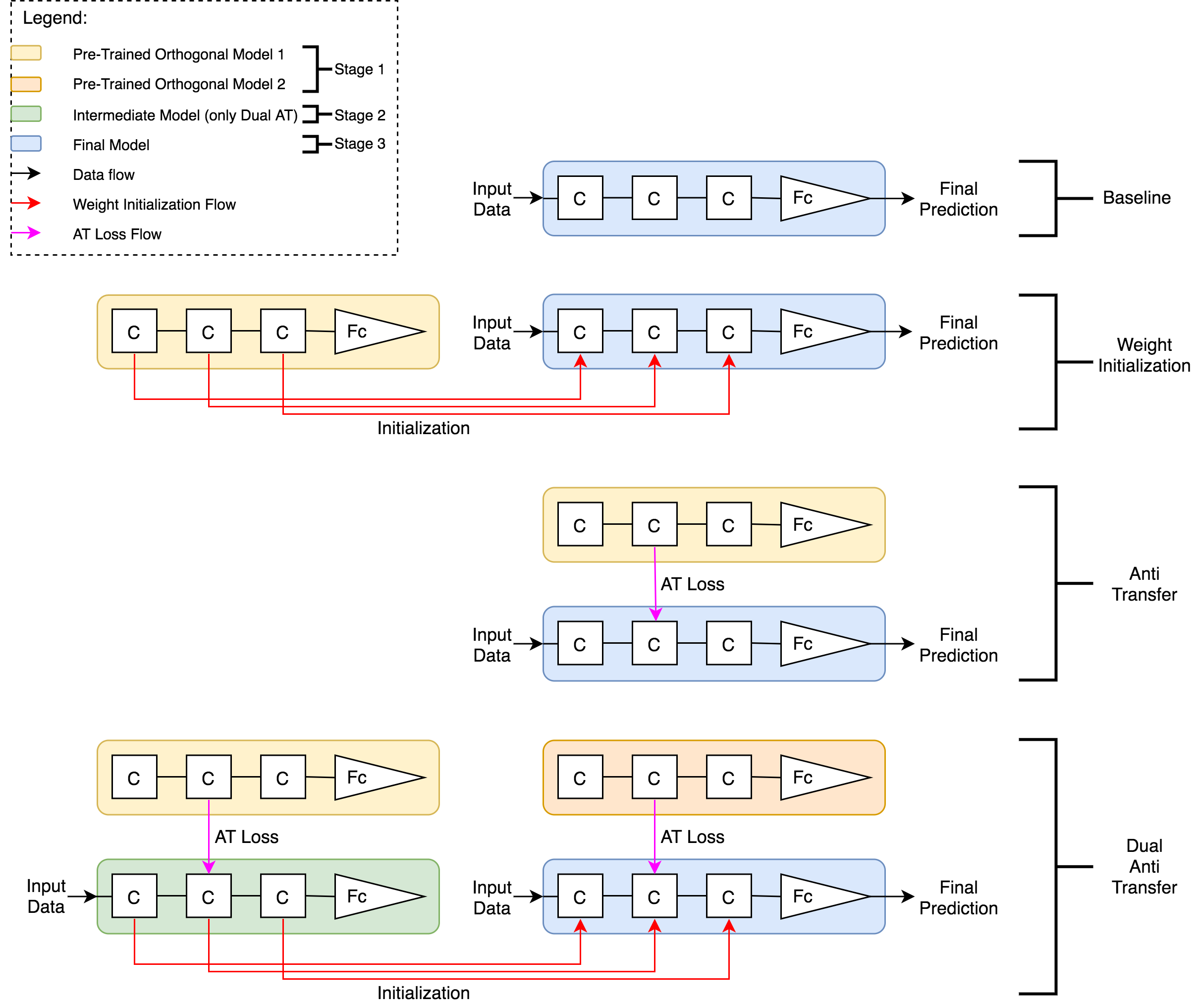}
\caption{Block diagram of our \textbf{training strategies}. The color coding reflects 3 consecutive temporal stages. 
Stage 1: pre-training of the orthogonal models (yellow and, only for dual AT, orange).  
Stage 2: only for dual AT, training of the intermediate model applying AT (green). 
Stage 3: training of the final models (blue) applying different transfer learning strategies: no transfer (baseline), weigh-initialization of the convolution layers, anti-transfer and dual anti-transfer. Different information flows are represented with differently colored arrows: the data flow is shown in black, the weight-initialization flow in red and the AT loss flow in magenta.}
\label{fig:exp_setup}
\end{figure}

We test anti-transfer learning on several audio classification tasks with 20 different combinations of training and pre-training  tasks in order to evaluate the behavior of anti-transfer learning in a variety of set-ups. 
We have three main classification tasks:  word recognition (WR), speech emotion recognition (SER) and sound goodness estimation (SGE) (i.e. how well musical notes are played by musicians \cite{romani_picas_oriol_2017_820937}). 

SER and SGE tasks are evaluated with two types of splitting the dataset into training, validation and test set: random split and class split by speaker or instrument. 
The class split types provide a more challenging task than the random split.
This is because these (orthogonal) classes reflect different data distributions in the random split the training, validation and test set distributions are the same. 
On the other hand, splitting by speaker or instrument presents a more realistic task for many applications.
The class-split is based on labels used in the orthogonal tasks (see Section \ref{subsec:dataset}).
This enables us to assess more directly the AT trained networks' invariance to the orthogonal task classes as discussed in Section \ref{subsec:discussion}.
For WR, we use only random split, but we added different types of background noise to the audio samples to create more challenging classification tasks. 
We test 3 scenarios: noise-free, low noise and high noise (see below for details) 

Our experiments are set up to test the effectiveness of anti-transfer learning, comparing it to the most common transfer learning method of weight initialization (WI) and to a baseline method without any transfer learning.
In this way we can compare anti-transfer to  regular transfer learning in the specific case of pre-training on orthogonal tasks.
In addition, we perform two further experiments (presented in Section \ref{subsec:ablation}).
In the first one we freeze the convolution layers in the WI modality up to the same layer where we apply the AT loss. 
This avoid possible dissipation of prior knowledge when  training.
In the second experiment we invert $\beta$ in the AT loss, so that  similarity of feature activations is encouraged instead of dissimilarity, i.e. performing the opposite of regular AT.
Figure \ref{fig:exp_setup} shows a diagram of the different training strategies we compared. 
We perform 3 consecutive training stages.  
First, we pre-train the models on the orthogonal tasks. Then,  only for dual AT, we apply AT to train an intermediate model  on  the  final  task. 
The weights of the intermediate model are then used to initialize the final model.   
Finally,  we  train our  final  models, applying the different transfer learning strategies.

\subsection{Datasets}
\label{subsec:dataset}

We use six different datasets overall. 
For our experiments, we extract subsets from larger datasets to reduce training times and adjust class imbalances. 
While this limits comparability to published results, it enabled us to perform a much broader range of experiments as reported in this and the following section. 
\begin{enumerate}

 \item \textit{Google Speech Commands}: A Dataset for Limited-Vocabulary Speech Recognition V2 (GSC) \cite{warden2018speech}. 
 Task: Single-word speech recognition.  ~2.7 hours of audio. 
 10 different words (digits) recorded by more that 2000 non-professional speakers in various acoustic environments.
   
 \item \textit{MS-SNSD}: The Microsoft Scalable Noisy Speech Dataset. A dataset and online subjective test framework \cite{reddy2019scalable}. Approximately 20 hours of audio. 
 Task: background noise type recognition. 
 11 different types of noise mixed with speech audio signals at volumes between -20 and -40 dBfs.

 \item \textit{Librispeech}: An ASR corpus based on public domain audio books \cite{Panayotov2015LibrispeechAA}. 
 Task: Single-word speech recognition. ~100 hours of audio, 40 speakers, 1000 single-word labels. 
 One-word excerpts from audio book recordings.
  
 \item \textit{IEMOCAP}: The  Interactive Emotional Dyadic Motion Capture Database \cite{busso2008iemocap}.  
 Tasks: speech emotion recognition, speaker recognition. ~7:30 hours of audio, 5 speakers, 4 emotion labels: neutral, angry, happy, sad. 
 Actors  perform semi-improvised or scripted scenarios on defined topics.
  
 \item \textit{Nsynth}: A large-scale, high-quality dataset of annotated musical sounds. \cite{nsynth2017}. Task: instrument eecognition. ~66 hours of audio. 
 11 different instrument macro-categories. 
 One-note recordings of musical instruments.
  
 \item \textit{Good-Sounds}: A dataset to explore the quality of instrumental sounds (GS) \cite{romani_picas_oriol_2017_820937}. 
 Tasks: sound goodness estimation, instrument recognition.  ~14 hours of audio. 12 different instruments, 5 different goodness rates. 
 One-note recordings of acoustic musical instruments, played by professional musicians.

\end{enumerate}
 
The above descriptions refer to the subsets we extracted (or generated, for MS-SNSD), not to the original size and arrangement of these datasets. 
Please refer to the references above for the original specifications.

For each target task, we pre-train on two different tasks for transfer and  anti-transfer learning. 
For word recognition we train on GSC and we pre-train on speech emotion recognition (IEMOCAP) and on background noise type recognition (MS-SNSD).
For speech emotion recognition we train on IEMOCAP and we pre-train on speaker recognition with the same training dataset (IEMOCAP) and on word recognition with a larger dataset (Librispeech). 
For sound goodness estimation we train on Good-Sounds, we pre-train on instrument recognition with the same training dataset (Good-Sounds) and with a larger dataset (Nsynth).

\subsection{Processing stages, training parameters and training strategies}

We paid particular attention to performimg all experiments (trainings and pre-trainings) in the  same conditions, in order to isolate the influence of anti-transfer and weight initialization in the results. 
All experiments are performed in a Python and PyTorch environment, using the VGG16 network architecture \cite{simonyan2014very}  (in the implementation from the \emph{torchvision} library\footnote{\url{https://pytorch.org/docs/stable/torchvision/models.html}}). 

We apply two architectural modifications to the standard implementation: we reduce the channel  number of the very first layer to 1 (since we use single-channel magnitude spectrograms) and we vary the number of output neurons to match the classes to the task.
 
We apply the same 
pre-processing to all datasets: 
\begin{enumerate}
    \item We first down-sample all audio data to 16KHz sampling rate. 
    \item Then we zero-pad/segment all sounds in order to have data-vectors of the same length for each task. We segment the audio as follows:
    \begin{itemize}
        \item In the word recognition target task, we use 1-second sound samples as provided in the GSC. 
        For the orthogonal noise classification task, we first generate ~20 hours of noisy speech from MS-SNSD and then we extract 1-second fragments with no overlap. 
        For emotion recognition, we extract 1-seconds fragments from IEMOCAP. 
        \item In the speech emotion recognition target task, we use 4-seconds sound samples from IEMOCAP.
        For the orthogonal task of word recognition, we extract segments containing only one word\footnote{We use  \url{https://github.com/bepierre/SpeechVGG} for this.} from Librispeech and then zero-pad them to 4-seconds.
        \item In the sound goodness recognition target and the orthogonal instrument recognition task, we use 6-second sounds, applying zero-padding to both Nsynth and Good-Sounds sounds. 
    \end{itemize}
    
    \item Only for GSC, we add noise to the segmented speech sounds at 3 different levels: no noise, low noise
    (-40 to -20 dBfs) and high noise (0 to -10 dBfs).
    The noise sounds are from the MS-SNSD datasets. 
    Like for MS-SNSD we use the MS-SNSD code\footnote{\url{https://github.com/microsoft/MS-SNSD}} to perform this operation. 

    \item Next we compute the Short-Time-Fourier-Transform (STFT) using 16 ms sliding windows with 50\% overlap, applying a Hamming window and discarding the phase information.
    \item Finally, we normalize the magnitude spectra of each dataset to zero mean and unit standard deviation, based on the training set's mean and standard deviation. 
\end{enumerate}

We perform all neural network trainings and pre-trainings with the same parameters. 
We use a learning rate of 0.0005, a batch size of 13 and the ADAM optimizer \citep{kingma2014adam}. 
We apply dropout at 50\% but neither $L_1$ nor $L_2$ regularization.
We randomly initialize the weights of all networks, except in the case of  weight initialization from a pre-trained network (for WI and dual AT).
We train for a maximum of 50 epochs and apply early stopping by testing at the validation loss improvement with a patience of 5 epochs.
We divide every dataset using subsets of approximately 70\% of the data for training, 20\% for validation and 10\% for the test set.
All of the above settings are kept constant for all datasets in all configurations: non-transfer, transfer, anti-transfer/dual anti-transfer and also for all pre-trainings.



These experiments are not designed to produce to state of the art results on these datasets, 
because we want to focus on the impact of anti-transfer learning.
Therefore we used specific subsets and we did not optimise network architectures and hyperparameters to the individual datasets in order to exclude any other sources of performance variation.

\subsection{Classification Results}\label{subsec:classification}

\begin{table*}[!tb]
\caption{Results of the \textbf{pre-training} in terms of classification accuracy. 
\emph{Classes} is the number of different class labels. 
\emph{Hours} describes the amount of recorded material in the subset that we used. 
The \emph{Train} and \emph{Test} columns contain the accuracy on the train and test sets.
  }
  \centering
  \vspace{.5mm}
  \begin{tabular}{l|l|r|r|r|r}
    \multicolumn{1}{c|}{\bftab{Dataset}}  & 
    \multicolumn{1}{c|}{\bftab{Task (Recognition)}}  & 
    \multicolumn{1}{c|}{\bftab{Classes}}  & 
    \multicolumn{1}{c|}{\bftab{Hours}}  & 
    \multicolumn{2}{c}{\bftab{Accuracy}}\\
    & & & &\bftab{Train} & \bftab{Test}
  \\ \hline \hline 
    Librispeech & Speech  & 1000 & 100 & 97.6 & 91.8 \\
    IEMOCAP (1 sec) & Speech Emotion  &4 &7.3 &85.6 &51.9 \\
    IEMOCAP (4 sec) & Speaker  &5 &7.3 & 99.8 & 96.5 \\
    Good-Sounds & Instrument & 12 & 14 & 100.0 & 100.0 \\
    Nsynth & Instrument & 11 & 66 & 98.1 & 69.9 \\
    MS-SNSD & Noise Type  &11 &20 &100.0 &99.8 \\
    \hline
    \hline
  \end{tabular}
  \label{table:pretraining}
\end{table*}

\begin{table*}[!tb]
  \caption{Accuracy results for the \textbf{word recognition (WR)} target task pm the Google Speech Commands (GSC) dataset with 3 levels of background noise added: None, Low and High.
  We pre-train on noise type recognition (Nse) with MS-SNDS dataset (MSS) and speech emotion recognition (Emo) with IEMOCAP dataset (IEC).
  We compare between no transfer learning (None), regular transfer learning by weight initialization (WI),
  anti-transfer (AT) and
  dual anti-transfer  (Dual AT, using two pre-training tasks). 
  The order of the pre-training tasks is shown in the second column.
  The best results per column are highlighted in bold font.}
  \centering
  \vspace{.5mm}

  \begin{tabular}{l|l|l|r|r|r|r|r|r}
  \multicolumn{1}{l|}{\bftab{Transfer}}  & 
  \multicolumn{2}{c|}{\bftab{Pre-training} }& 
  \multicolumn{3}{c|}{
  $\vcenter{
  \hbox{\strut \bftab{Train accuracy}}
  \hbox{\strut \bftab{Noise level}}}$
  }
  &
  \multicolumn{3}{c}{
    $\vcenter{
  \hbox{\strut \bftab{Test accuracy}}
  \hbox{\strut \bftab{Noise level}}}$
  }\\
    \bftab{Type} 
    & \bftab{Task}
    & \bftab {Data}
    & \bftab{None} 
    & \bftab{Low} 
    & \bftab{High} 
    & \bftab{None} 
    & \bftab{Low} 
    & \bftab{High}  
    
    \\ \hline \hline
    None &n/a &n/a &98.45 &97.94 &97.23 &95.32 &93.67 &90.44\\
    \hline
    WI & Noise  & MSS &98.33 &97.85 &96.34  &94.83 &93.97 &90.51\\
    WI & Emo & IEC &98.67 &97.69 &97.36  &95.40 &93.51  &90.35 \\
    \hline
    \bftab AT & Noise  & MSS &99.57 &99.11 &98.42  &95.70 &94.81 &90.99\\
    \bftab AT & Emo  & IEC &99.02 &99.09 &98.36  &95.57 &94.91 &\bftab{91.38}\\
    \hline
    Dual \bftab AT
    & $\vcenter{\hbox{\strut {Emo +}}\hbox{\strut {Nse }}}$ 
    & $\vcenter{\hbox{\strut {IEC +}}  \hbox{\strut {MSS}}}$ 
    & \bftab 99.84 &\bftab 99.49 & 98.29  &\bftab 96.60 &94.91 &90.98
    \\
    Dual \bftab  AT
    & $\vcenter{\hbox{\strut {Nse +} }\hbox{\strut {Emo }}}$
    & $\vcenter{\hbox{\strut MSS 
    +}\hbox{\strut IEC}} $
     & 99.31 &99.17 &\bftab{98.89}  &95.64 &\bftab{95.20} &90.67
    \\ \hline \hline
  \end{tabular}
  \label{table:results_digits}
\end{table*}
\begin{table*}[tb]
  \caption{Accuracy results for the \textbf{speech emotion recognition (SER)} target task on the IEMOCAP dataset. 
  Comparison between no transfer learning (None), weight initialization (WI)
  and anti-transfer (AT) with pre-training on different datasets. 
  In particular, we compared anti-transfer with pre-training on the same dataset (IEMOCAP) but on an orthogonal task (speaker recognition) and on a bigger dataset (Librispeech) on a different orthogonal task (word regognition). We test 2 different train/validation/test split: random (Rand) and speaker-wise (Speaker).
  The best results per column are highlighted in bold font.}
  \centering
  \vspace{.5mm}

  \begin{tabular}{l|l|l|r|r|r|r}
  \multicolumn{1}{l|}{\bftab {Transfer}}  & \multicolumn{2}{c|}{\bftab {Pre-training}} & 
  
  \multicolumn{2}{c|}{  
    $\vcenter{
  \hbox{\strut \bftab{Train accuracy}}
  \hbox{\strut \bftab{Split Type}}}$}&
  \multicolumn{2}{c}{  
    $\vcenter{
  \hbox{\strut \bftab{Test accuracy}}
  \hbox{\strut \bftab{Split Type}}}$} 
  \\
    \bftab {Type} 
    & \bftab {Task} 
    & \bftab {Dataset} 
    & \bftab {Rand}  
    & \bftab {Speaker} 
    & \bftab {Rand}
    &\bftab {Speaker} 
    \\ \hline \hline
    None & n/a & n/a & 69.0 &67.8 &63.7 &57.2\\
    \hline
    WI
     & Word  & Librispeech  &66.9 &66.9  &63.4 &59.2\\
    WI  & Speaker  & IEMOCAP  &70.7 &66.9  &64.8 &58.5\\
    \hline
    \bftab {AT} & Word & Librispeech  &72.0 &68.6  &\bftab {66.9} & 61.1\\
    \bftab {AT} & Speaker & IEMOCAP  &\bftab {75.5} &\bftab {74.5}  &66.5 & \bftab {61.3}\\
    \hline
    \hline
  \end{tabular}
  \label{table:results_emotion}
\end{table*}
\begin{table*}[tb]
  \caption{Accuracy results for \textbf{sound goodness estimation (SGE)}. For the target task we use the Good-Sounds dataset. We compare no transfer learning (None), weight initialization (WI)
  and anti-transfer (AT). 
  In particular, we compare anti-transfer with pre-training on the same dataset (Good-Sounds) and on a bigger dataset (Nsynth). We test 2 different train/validation/test splits: random (Rand) and instrument-wise (Instr).
  The best results per column are highlighted in bold font.}
  \centering
  \vspace{.5mm}

  \begin{tabular}{l|l|l|r|r|r|r}
  \multicolumn{1}{l|}{\bftab{Transfer}}  & \multicolumn{2}{c|}{\bftab{Pre-training}} & 
  
  \multicolumn{2}{c|}{  
    $\vcenter{
  \hbox{\strut \bftab{Train accuracy}}
  \hbox{\strut \bftab{Split Type}}}$}&
  \multicolumn{2}{c}{  
    $\vcenter{
  \hbox{\strut \bftab{Test accuracy}}
  \hbox{\strut \bftab{Split Type}}}$} 
  \\
    \bftab{Type}
    & \bftab{Task}
    & \bftab{Dataset}
    & \bftab{Rand} 
    & \bftab{Instr}
    & \bftab{Rand}
    & \bftab{Instr} 
    \\ \hline \hline
    None  &n/a &n/a &91.8 &42.2 &83.8 &22.8\\
    \hline
    WI & Instrument & Nsynth &93.4 &40.5  &84.7 &29.6\\
    WI & Instrument  &Good-Sounds &93.3 &\bftab{42.3}  &84.9 &23.9\\
    \hline
    \bftab{AT} &Instrument  & Nsynth &\bftab{96.8} &41.0  &\bftab{86.3} & 30.0\\
    \bftab{AT} &Instrument  & Good-Sounds &93.9 &36.4  &85.7 &\bftab{34.3}\\
    \hline
    \hline
  \end{tabular}
  \label{table:results_goodness}
\end{table*}

Table \ref{table:pretraining} shows the results of the pre-training in terms of classification accuracy. 
There is wide variation in performance on the different tasks, with the Good-Sounds and MS-SNSD saturating or almost saturating on the train and test set for instrument and background noise type recognition.

Tables \ref{table:results_digits}, \ref{table:results_emotion} and \ref{table:results_goodness} show the results obtained on the target tasks of word recognition, speech emotion recognition and sound goodness estimation, respectively.
These tables contain the baseline results without transfer learning (None), with  standard transfer learning using weight initialization (WI) and with anti-transfer learning (AT) on 20 pre-task/actual-task combinations in total. 
While for SER and SGE we test only anti-transfer with one orthogonal task at a time, for WR we additionally test dual anti-transfer (Dual AT), applying two orthogonal tasks as described in Section \ref{method}. 

We applied anti-transfer to one layer of the VGG16 network with each of the 13 convolution layers for each task.
The reported anti-transfer test accuracy results reflect the choice of layer that reached the best validation accuracy.
In all experiments, the coefficient $\beta$ is fixed to 1 since, as further analyzed in Section \ref{subsec:beta}, this provides the best accuracy results.

The results 
(Tables \ref{table:results_digits}, \ref{table:results_emotion} and \ref{table:results_goodness}) show that anti-transfer improves the test accuracy in all cases and interestingly improves also the training accuracy in all cases but one (sound goodness estimation with instrument-wise split dataset, Table \ref{table:results_goodness}), compared to both the baseline and weight initialization.
We have a maximum improvement in the test accuracy of 11.5 percentage points (pp) (for sound goodness estimation with instrument-wise split dataset, Table \ref{table:results_goodness}) and a maximum improvement in the training accuracy of 6.7 pp (for speech emotion recognition with speaker-wise split dataset, Table \ref{table:results_emotion}).
The overall average improvement is of 4.11 pp for the test accuracy and of 2.35 pp for the training accuracy.
\begin{figure}[tb]
  \centering
  \includegraphics[width=12.8cm]{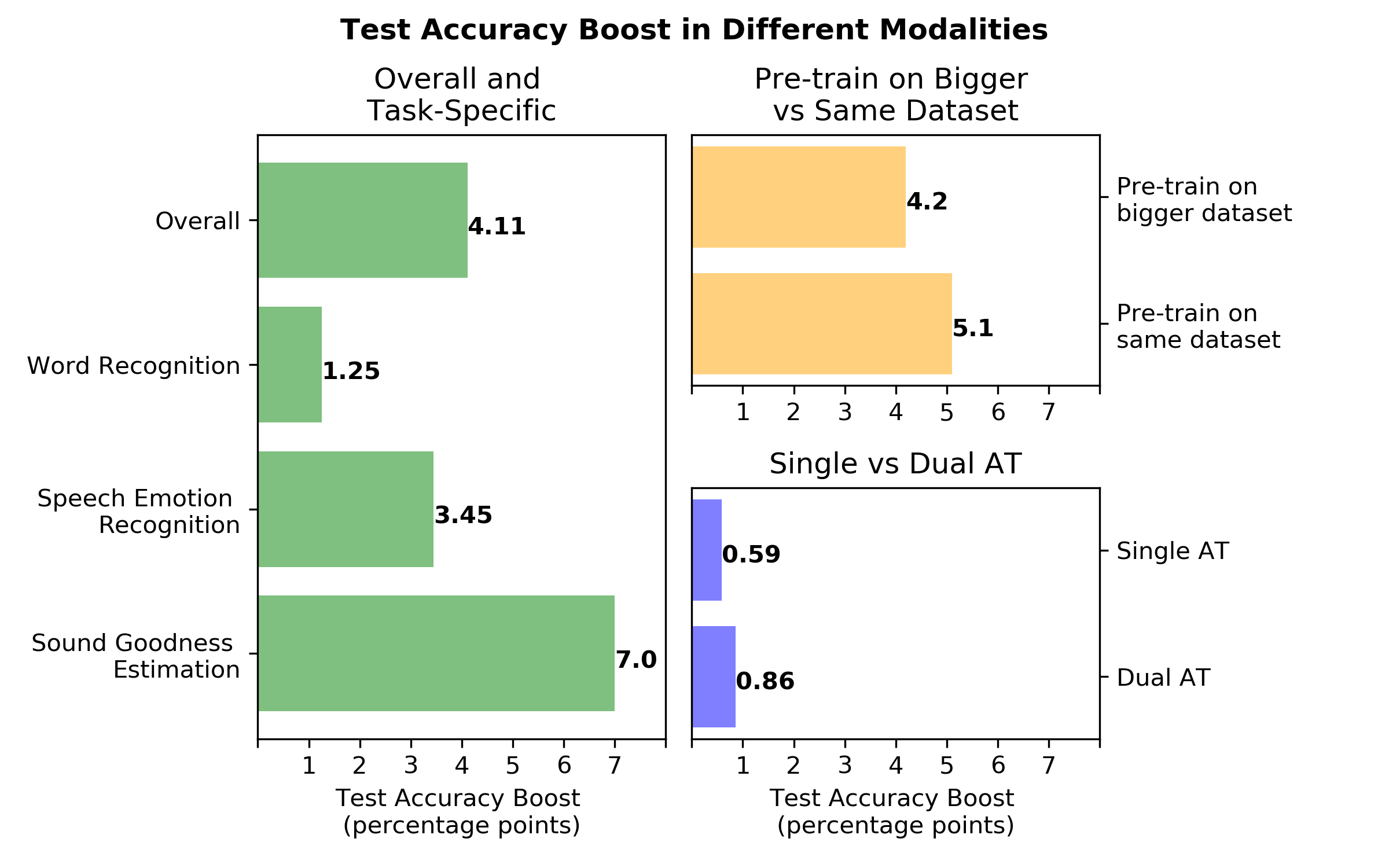}

\caption{Average \textbf{improvement} by applying anti-transfer learning on different applications and different settings compared to the baseline (no transfer learning). The overall and the task-specific measures (word recognition, speech emotion recognition, sound goodness estimation, in green) show the average over the best improvements on each task/split or task/noise level configuration. 
The other measures 
show the average improvement over all experiments of a modality.
The pre-training on bigger/same dataset modality (orange lines) is computed for Good-Sounds per-trained on NSynth and itself and for IEMOCAP pre-trained on Librispeech and itself. 
The single/dual AT modality (blue lines) is computed for the Google Speech Commands dataset pre-trained on MS-SNSD and IEMOCAP.
}
\label{fig:boost}
\end{figure}
Figure \ref{fig:boost} shows the average gain achieved by anti-transfer learning in the test accuracy for different tasks and settings. 
It has practical relevance that the improvement in the networks' generalization is higher when anti-transfer is applied with a feature extractor trained on an orthogonal task with the same dataset as opposed to a different but larger dataset (we tested this property only on SER and SGE:  IEMOCAP pre-trained on speaker recognition vs IEMOCAP trained on speech emotion recognition and Good-Sounds pre-trained on instrument recognition vs Good-Sounds trained on sound goodness estimation).

Another interesting aspect is that using dual anti-transfer provides a higher accuracy boost compared to anti-transfer on a single orthogonal task (we tested this only on WR: GSG pre-trained on speech emotion recognition and background noise type recognition).
This suggests that the task invariance effect of anti-transfer learning can be cumulative, opening the possibility of pre-training on multiple orthogonal tasks.

\section{Analysis and Discussion}\label{sec:analysis}

The results in the previous section show a robust improvement resulting from the use of anti-transfer learning. 
Here we investigate various aspects of the method for understanding and optimizing its performance. 

\subsection{Ablation Study: Encouraging Similarity vs. Dissimilarity}\label{subsec:ablation}
\begin{figure}[tb]
  \centering
  \includegraphics[width=11cm]{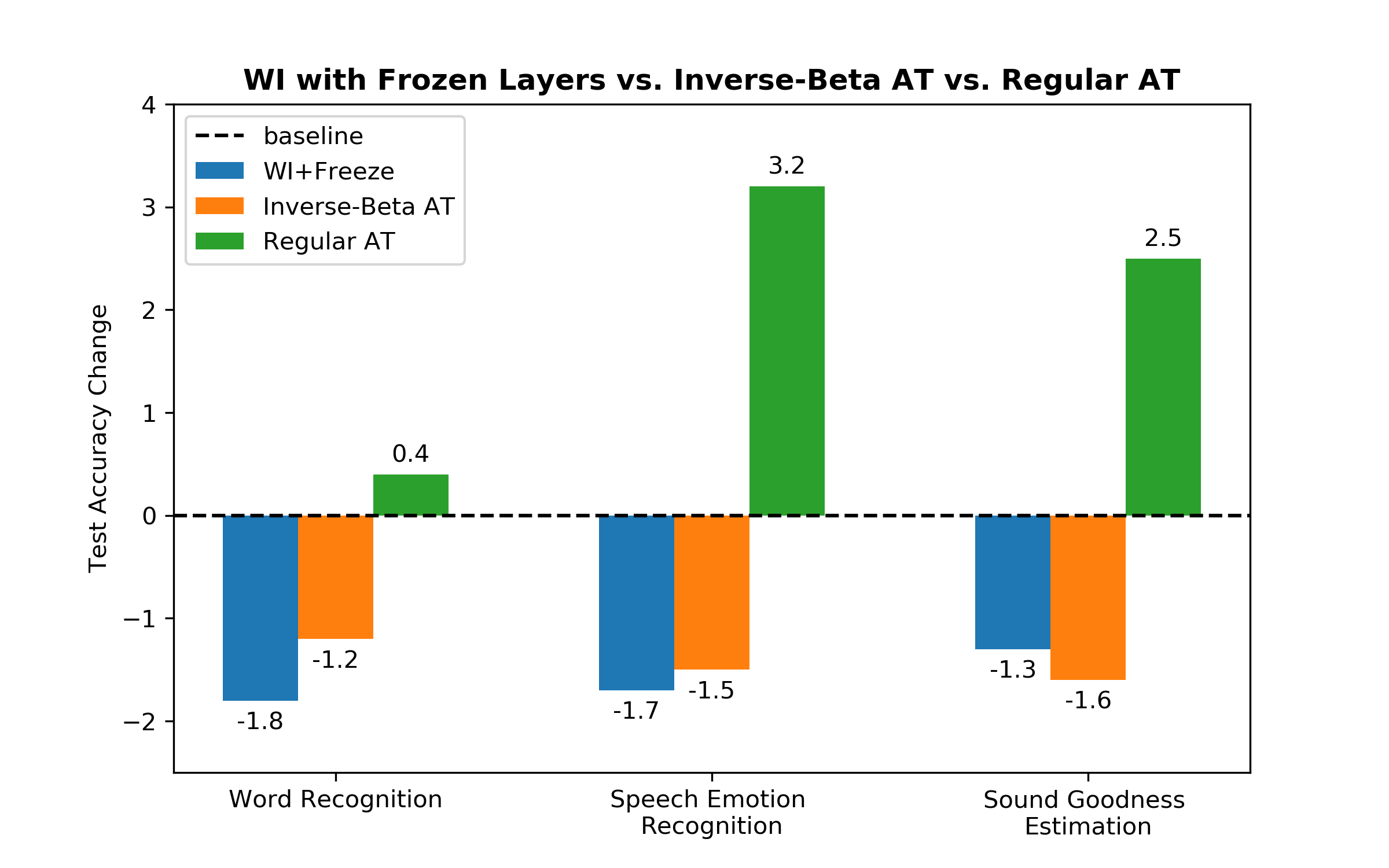}
\caption{Comparison of \textbf{regular AT} (encouraging feature dissimilarity with the orthogonal task, green columns), \textbf{inverse-beta AT} (encouraging feature similarity, orange columns) and \textbf{weight initialization} with \textbf{frozen} convolution layers (until the same layer where we apply AT, blue columns) on all target tasks: word recognition, speech emotion recognition and sound goodness sstimation. 
The improvement in the test accuracy is shown, comparing to the baseline results (no AT nor WI applied, black segmented line: 95.3\% for word recognition, 63.7\% for speech emotion recognition, 83.8\% for sound goodness estimation).
}
\label{fig:inverse_beta}
\end{figure}
As an ablation study, we performed additional experiments where we encourage the models to develop representations that are similar instead of dissimilar to the models pre-trained on orthogonal tasks. 
The results are shown in Figure~\ref{fig:inverse_beta}. 
We tested two methods for encouraging feature similarity.
The first consists of inverting the sign of the $\beta$ parameter to encourage similarity instead of dissimilarity through the AT loss. 
This operation can be considered as the opposite of the regular AT (in line with \cite{beckmann2019speech}).
The second
consists of weight initialization and freezing (i.e. setting as not trainable) all convolution layers  from the input layer of the network up to the layer where we apply AT.
This test is complementary to the comparison between AT and WI, since in regular WI the knowledge transferred from the pre-trained model may completely disappear during the training because of the \textit{catastrophic forgetting} phenomenon \cite{DBLP:journals/nn/ParisiKPKW19}.
This experiment shows the model's performance when we 
avoid this phenomenon by freezing the initial layers. 
We performed these two experiments 
using the task/orthogonal-task/AT-layer combination that yielded the best performance in each case, which are:
\begin{itemize}
    \item Word recognition: GSC with no further noise added, background noise type recognition pre-training (MS-SNSD), layer 5.
    \item Speech emotion recognition: IEMOCAP random split, word recognition pre-training (Librispeech), layer 5.
    \item Sound goodness estimation: GS random split, instrument recognition pre-training (NSynth), layer 6.
\end{itemize}


The results show that both inverse-Beta-AT and freeze-WI configurations lead to a decreased performance compared to regular AT and to the baseline (no transfer learning).
These results support the motivating idea of anti-transfer learning: given a suitable choice of orthogonal tasks, avoiding similar representations can improve learning and generalization on the target task.
Conversely, while transfer learning has proven efficient and effective in many settings, for orthogonal tasks like in our experiments it can actually be detrimental. 


\subsection{Convolutional Feature Activations}
\begin{figure}[tb!]
  \centering
\centerline{
  \includegraphics[width=14.5cm]{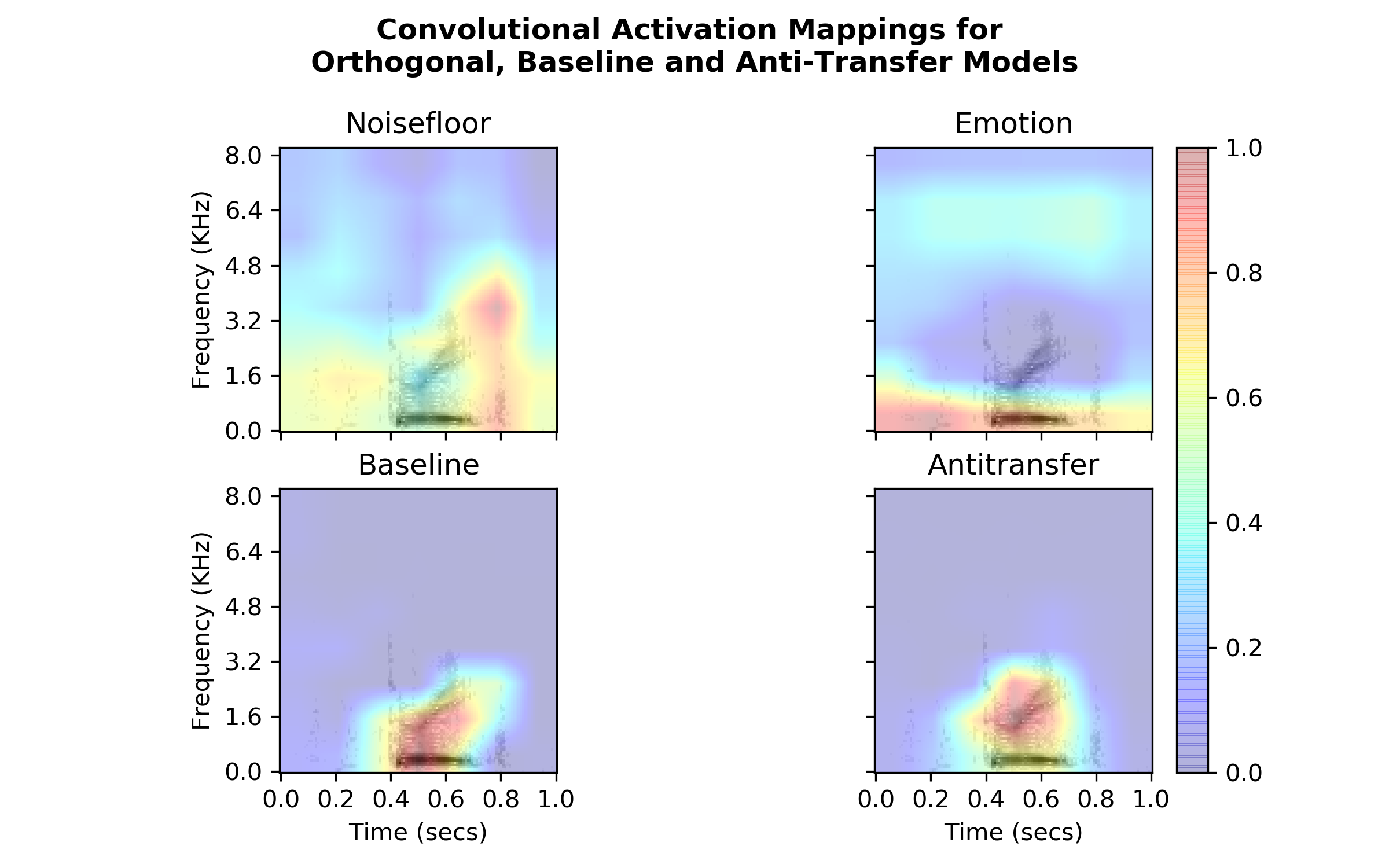}
}
  
\caption{Grad-CAM convolutional \textbf{feature activations} of different models for the same input data-point from the test set of the GSC with low noise added (word recognition target task). 
The activations have been computed in the last (13th) convolution layer, where we applied anti-transfer for this experiment.
In all plots, the magnitude spectra are shown in black.
The top row shows: the activation of the model trained for the first orthogonal task (noise type recognition, left), the activation of the model trained for the second orthogonal task (emotion recognition, right). 
The bottom row shows: the activation of the baseline model (no transfer or anti-transfer, left), the activation of the dual anti-transfer model (pre-trained on noise type and emotion recognition, right)}
\label{fig:activations}
\end{figure}
In order to support a visual interpretation of the deep representations generated with anti-transfer learning, we applied the Grad-CAM technique \cite{DBLP:journals/ijcv/SelvarajuCDVPB20} to our trained models.\footnote{We used a modified version of \url{https://github.com/jacobgil/pytorch-grad-cam} .} 
In a CNN, Grad-CAM produces class-discriminative localization maps of a convolution layer using the gradient of the classification score with respect to the convolutional features present in that layer.
This produces a heatmap of the same dimension as the input data, showing which parts of the input matrix are most important for classification.
Please refer to the above mentioned paper \cite{DBLP:journals/ijcv/SelvarajuCDVPB20} for an in-depth description of this technique.

For this visualization we used the GSC dataset with low noise added, where we apply dual AT.
We selected this specific case to better assess the effectiveness of our approach in moving away from unwanted features, showing the behavior of AT with 2 simultaneous orthogonal tasks.
Figure \ref{fig:activations} shows the Grad-CAM activations obtained for a datapoint of the test set, containing a male voice saying the world ``eight" with added ``office-like" background noises at low volume. 
The voice appears as a in the center of the lower half of the spectrogram (approximately from 0.4 until 0.7 secs), while the background noise appears mainly as vertical spikes outside of the center  (approximately at 0.12, 0.22, 0.38, 0.8 secs).
The activations shown are obtained for the two models trained on the orthogonal tasks (background noise recognition and emotion recognition), the baseline model (no transfer learning) and the dual AT model with AT applied on the last convolution layer (pre-trained first on background noise recognition and then on emotion recognition).
As expected, the background noise type recognition model focuses mostly on pixels outside the center, in particular on the spike at 0.8 secs. 
The emotion recognition model focuses instead mostly on the lower frequencies in the spectrum (approximately below 800 Hz), which is the normal range for the fundamental frequency of the human voice.
The baseline model successfully focused on the speech signal in the center, although it slightly expands also towards the noise spike at 0.8 secs and it has a high activation in the low-frequency region where emotion information is more present (according both to our orthogonal model and our research experience).
Similarly, the dual AT model is focused on the speech signal center, but it adjusted its attention towards the mid fequencies, where most format and consonant information is present, decreasing its activation on both the low-frequency area (emotion) and the spike at 0.8 secs (background noise).
This example confirms that the dual AT model developed a certain degree of invariance to both orthogonal tasks (noise type and emotion recognition) when predicting the between the target task (word recognition), which underpins the observed effectiveness of anti-transfer learning in our experiments.


\subsection{Layer Selection}
\label{subsec:layer_selection}
\begin{figure}[tb]
  \centering
  \includegraphics[width=11cm]{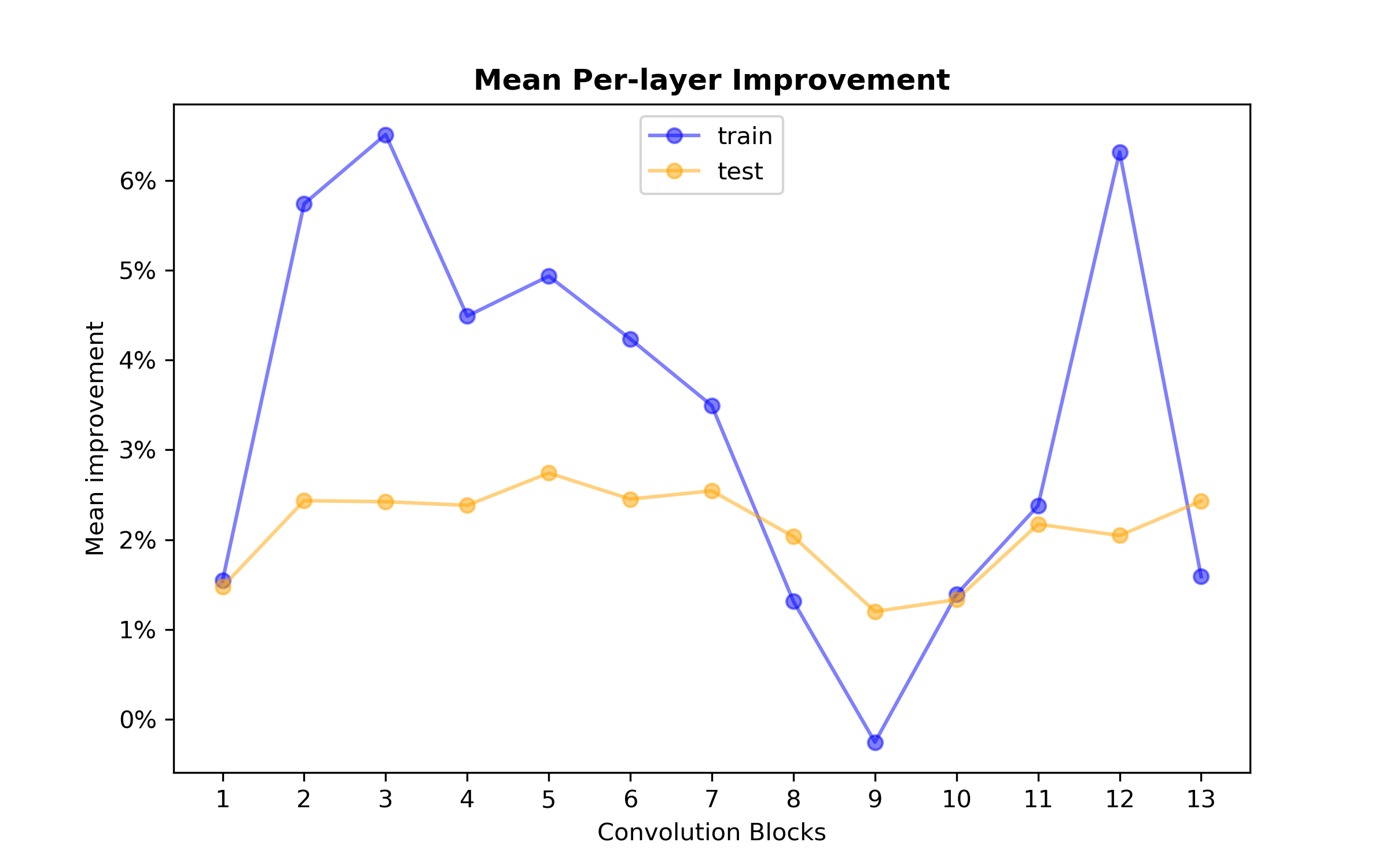}
\caption{Mean \textbf{per-layer improvement} on speech emotion recognition (IEMOCAP random split) with pre-training on word recognition (Librispeech). The improvement refers to the baseline with no weight initialization.}
\label{fig:layer_acc}
\end{figure}
We tested all layers in all task/orthogonal task combinations and Figure~\ref{fig:layer_acc} shows the average per-layer improvement in both train and test accuracy that we obtained in the speech emotion recognition task.
In this case, computing the anti-transfer loss with layer 5 provides the best performance, although layers 7 and 13 yield comparable results.
Moreover, in both training and test, layer 9 yields the lowest performance and it is the only one that leads to a slight training accuracy decrease.
However, most other layers also lead to improvements and the situation may vary when using different  architectures or datasets.
Also for word recognition layer 5 yields the best results, but for sound goodness estimation we obtained the best performance with layer 6. 

In summary, there is no overall unequivocal best choice for the layer to use for the anti-transfer loss. 
Our intuitive expectation was the last layers would be most effective, as they should be most task-specific according to \cite{yosinski2014transferable}. 
It is interesting to observe that these results of are not reflected in our layer-wise evaluation, but we don't currently have an explanation for this. 

Based on these results, we experimented with training using the anti-transfer loss on multiple layers at the same time.
We tried to use three layers at once in three configurations: the first convolution layers, the last ones, the best ones according to the results above. 
These configurations yielded worse results than the baseline setting (non-transfer learning).
However, this may be because we did not perform parameter optimization on this approach, therefore further exploration could potentially lead to positive results. 

\subsection{Learning dynamics}
\begin{figure}[tb!]
  \centering
  \includegraphics[width=11cm]{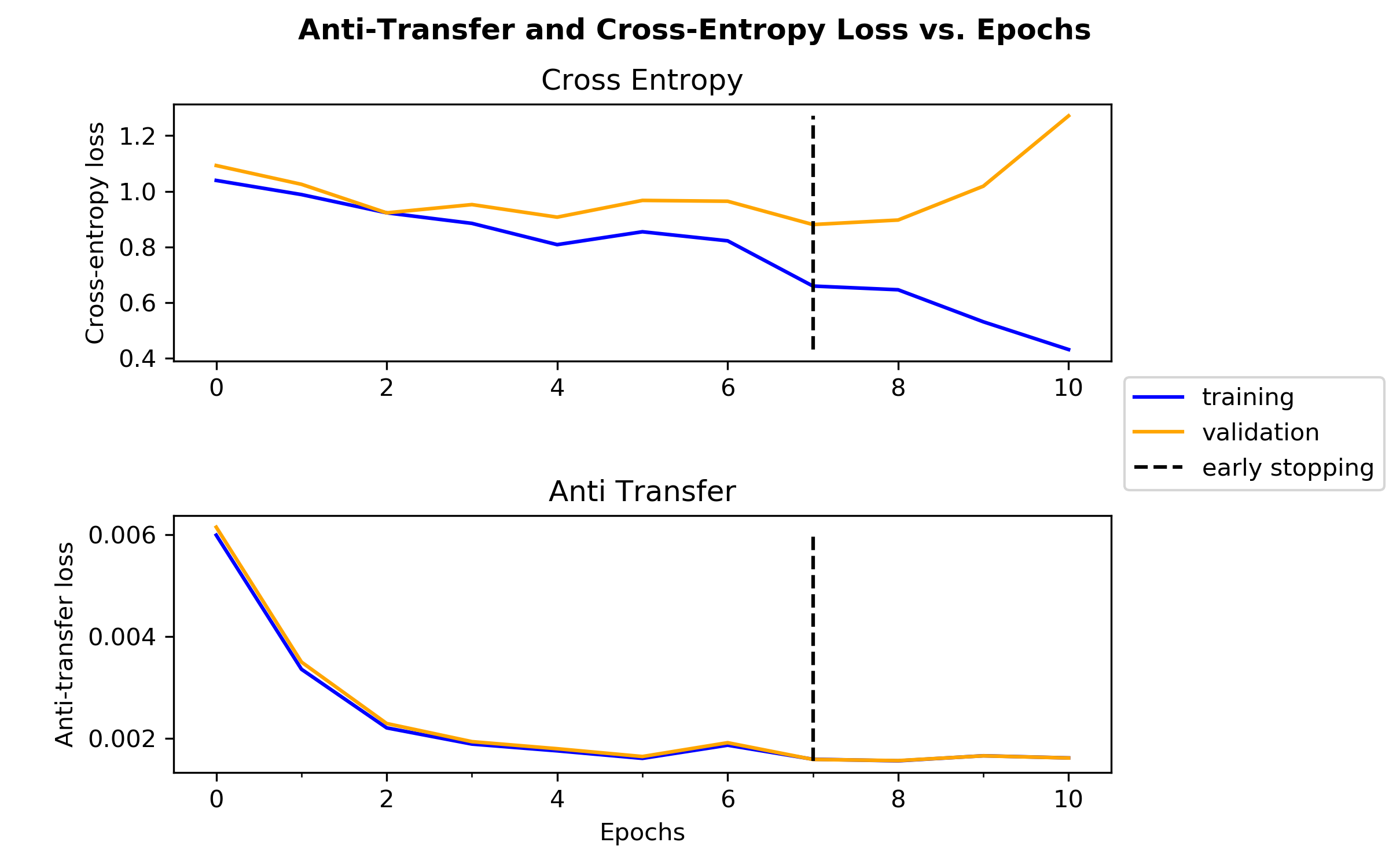}
\caption{Evolution of the train and validation cross-entropy loss and train and validation anti-transfer \textbf{loss} during the training. 
This example refers to training on speech emotion recognition as target task (IEMOCAP random-split) and pre-training on word recognition (Librispeech) and anti-transfer applied to the 5th convolution layer.}
\label{fig:loss_epochs}
\end{figure}
Figure \ref{fig:loss_epochs} shows the development of the classification loss (cross-entropy) and the anti-transfer loss during training for speech emotion recognition (IEMOCAP random-split), with pre-training on word recognition (Librispeech) and anti-transfer applied to the 5th convolution layer. 
Here it is evident that the network is actually learning to differentiate its deep representations from the pre-trained ones, as the anti-transfer loss is substantially reduced. 
Moreover, as we expected, the anti-transfer loss is already low from the first epoch because the randomly initialized feature maps start mostly uncorrelated to the ones of the pre-trained network. 
The relatively low magnitude of the anti-transfer loss with respect to the cross-entropy loss indicates that anti-transfer plays a ``preventive'' 
role during training, keeping the deep representations from becoming correlated.

\begin{table*}[tb]
  \caption{Accuracy results for different \textbf{channel aggregation} methods and different \textbf{similarity functions}. 
  All results are computed for speech emotion recognition as target task (IEMOCAP random-split) with pre-training on word recognition (Librispeech) and anti-transfer on the 5th convolution layer. 
  The best training, validation and test accuracy results overall are highlighted in bold font.}
  \centering
  \vspace{.5mm}

  \begin{tabular}{l|c|c|c|c|c|c}
  \multicolumn{1}{l|}{}  &\multicolumn{3}{c|}{\bftab {Sigmoid MSE Similarity}} & \multicolumn{3}{c}{\bftab {Squared Cos Similarity}}\\
    \bftab {Aggregation} &\bftab {Train} &\bftab {Val} &\bftab {Test} &\bftab {Train} &\bftab {Val} &\bftab {Test}
    \\ \hline \hline

    Mean &68.2 &68.3 &63.9 &68.7 &66.1 &60.0\\
    Sum &68.4 &68.2 &63.8 &69.5 &66.0 &60.0\\
    Comp Mul &71.0 &67.5 &63.9 &70.4 &67.0 &63.1\\
    Max &68.3 &66.3 &65.0 &\bftab {76.7} &66.2 &66.7\\
    \bftab{Gram Matrix} &76.3 &65.9 &65.8 &72.5 &\bftab {68.7} &\bftab {66.9}\\

    \hline
    \hline
  \end{tabular}

  \label{table:gridsearch}
\end{table*}

\subsection{Aggregation and Distance Functions}
Table \ref{table:gridsearch} shows the results of experiments performed to select the best channel aggregation and similarity function to compute the anti-transfer loss.
All aggregation types refer to a function applied pixel-by-pixel along the channel dimension. \textit{Comp Mul} stands for compressed multiplication (feature activation values raised to the power of 0.001 and then multiplied along the channel dimension). 
The compression is necessary when multiplying pixel-by-pixel to avoid rounding to 0 during the multiplication of many small numbers.

As an alternative to Squared Cosine Similarity we used \textit{Sigmoid MSE Similarity}, which we define as the negative standard Mean Squared Error with a sigmoid function applied to avoid excessive loss values. 
Without the sigmoid, the training led to very high absolute values in the feature maps, which minimizes the AT loss, but also drastically decreased the accuracy. 
We also tried several approaches to compute the similarity for all possible channel combinations without using any aggregation method, but all of them were too expensive 
in terms of computation or memory.
We find that to aggregate the channel information using the Gram matrix and to compute the matrix similarity with squared cosine similarity gives the best results, which is why we used this combination 
in the experiments in the previous Section~\ref{sec:experimental}.

\subsection{AT Loss Weight}
\label{subsec:beta}

\begin{figure}[tb]
  \centering
  \includegraphics[width=11cm]{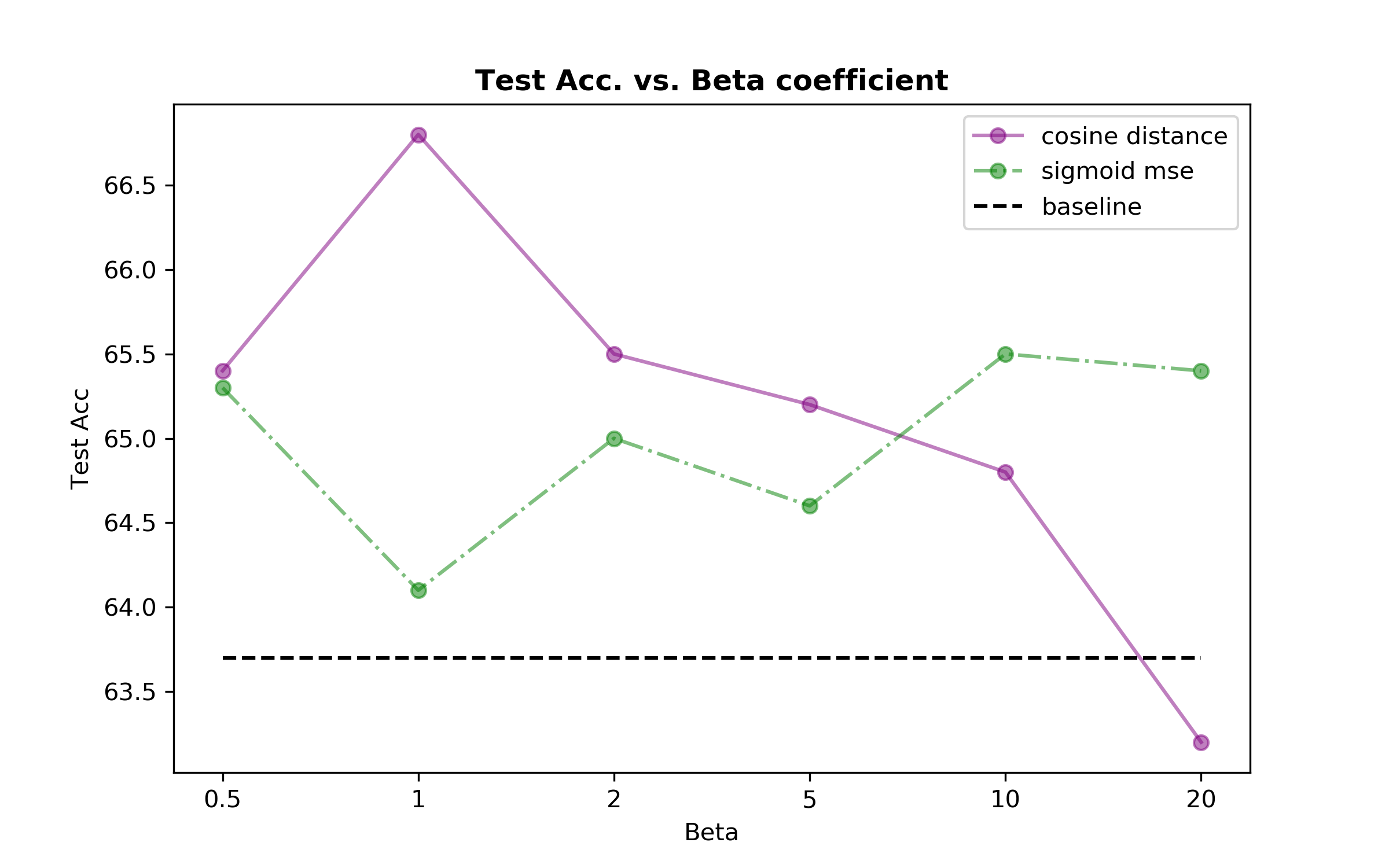}
\caption{Variation of the test accuracy for \textbf{different \boldmath$\beta$ parameters} (weight of the AT loss) using Gram aggregation with the squared cosine similarity (solid line) and the sigmoid MSE (dash-dotted line). 
This  example  refers  to  training  on  speech emotion recognition as target task (IEMOCAP random-split), with pre-training on word recognition (Librispeech) and anti-transfer applied to the 5th convolution layer (the one yielding the best result). 
}
\label{fig:loss_beta}
\end{figure}
The $\beta$ parameter that determines the weight of the anti-transfer loss  has a clear impact on the performance. 
As shown in Figure~\ref{fig:loss_beta} for emotion recognition (IEMOCAP  with random splitting) with pre-training on Librispeech and anti-transfer applied to the 5th convolution layer,  we get the best result using the squared cosine similarity with a $\beta$ value of 1, obtaining a performance gain of approximately 3 percentage points. The performance gain is smaller for all other values and there is no improvement with a very high (20) $\beta$. 
For sigmoid MSE the performance gain is smaller, but less dependent on the $\beta$ value. 
For practical purposes, $\beta = 1$ with cosine similarity seems to be a good default choice. 

\subsection{Computation and Memory Costs}
The improved accuracy comes at a cost of increased computational and memory demands at training time. 
The following considerations refer to our specific implementation and different strategies may different trade-offs. 
For instance, it is possible to pre-compute all needed Gram matrices in advance and avoid loading the pre-trained feature extractor into the GPU RAM. 
This would lead to lower memory demand and computation time during training, but it would be incompatible with in-place data augmentation and other online approaches.

With reference to our implementation, regarding computational time, training a network with anti-transfer learning takes on average approximately $2.8$ times longer compared to the same network with standard training. 
This refers to the only training with anti-transfer applied, without taking into account the time needed to pre-train the feature extractor on the orthogonal task. 
Moreover, learning with AT loss requires more memory than standard training, since it requires to fit into memory the trained network, the pre-trained feature extractor, the feature maps and the Gram matrices to be compared. 
The size of these depends on the chosen architecture, the input data dimension and the the number of channels of the convolution layer(s) used to compute the anti-transfer loss.
The additional memory $M_t$ required to compute the anti-transfer loss using a network pre-trained on an orthogonal task $t$ can be calculated as:
\begin{equation}
    M_t = E_t +  \sum_{l \in L} 2 (\#G_l + \#F_l) \times bytes\_per\_number
\end{equation}
where $E_t$ is the size of convolutional part of the pre-trained network,  $L$ is the set of layers used for anti-transfer, $\#G_l$ is the size of the Gram matrix computed on the layer $l$ , $\#F_l$ is the size of the feature map of layer $l$. 
The term inside the summation is multiplied by $2$ because we compute the above-described matrices both for the currently-trained and the pre-trained network.
The term $\#G_l$ is determined by \(batchSize \times numChannels^2\), while $\#F_l$ depends on all dimensions of the input data, on the network's architecture and on the layer parameters. 
The $bytes\_per\_number$ is 4 in our case.
In our specific test case with the VGG16 network, the whole network occupies $\sim$1620 MB, while the feature extractor $E_t$ requires additional $\sim$1150 MB,
The dimension of one batch with one single GSC data point pre-processed as described above is $[1,1,126,129]$. 
With this configuration the term \(G_l + F_l\) is $\sim$62 MB when anti-transfer is computed only on the first layer (shape [1,64,126,129]) and is $\sim$714 MB when computed on the last layer (shape [1,512,7,8]). 

\subsection{Discussion}\label{subsec:discussion}

\paragraph{Results}

Anti-transfer leads to a robust improvement in test results in all our experiments. 
The learning dynamics, data splits and the visualization show that the similarity between the pre-trained and the new network's representations is effectively reduced. 
It seems that avoiding features from orthogonal tasks is generally helpful. 
The improvement with anti-transfer is generally greater when the baseline accuracy is lower.

Training results are also improved in most cases. 
This is unexpected, as we assumed that learning from scratch would find a good fit for the training set and that anti-transfer would only benefit generalization, as in regularization. 
However, it seems that for suitably chosen orthogonal tasks avoiding the representation of the pre-trained network not only avoids fitting to confounding aspects of the data, but even leads to a better fit to the target task during training. 
This contravenes the common assumption that end-to-end learning with deep learning leads to a near-optimal fit to the training data.
Instead it shows that the use of prior knowledge, here in the form of an orthogonal task, can help not just to improve generalization. 


In some cases, the train/test split was separating classes that the network was aiming to recognise in the orthogonal pre-training task (speaker-wise split for emotion recognition vs speaker recognition, Table \ref{table:results_emotion}, and instrument-wise split for sound goodness estimation vs instrument recognition, Table \ref{table:results_goodness}).  
When the orthogonal task was speaker or instrument recognition, we observed a significantly improved generalization to unseen speakers in the speaker-wise split and to new musical instruments in the instrument-wise split, respectively. 
This indicates that the models are actually developing a degree of invariance to the orthogonal tasks, which is also illustrated in the visualization example (Figure~\ref{fig:activations}).

The results show that pre-training on the same dataset provides higher improvement on average, compared to pre-training on a different dataset, even a much bigger one (Figure \ref{fig:boost}).
It is surprising that the larger dataset does not have a more positive effect.
We hypothesize that a more specific separation of representations can be developed for the specific  dataset with orthogonal task labels on the same dataset by more directly modelling the interactions between different tasks. 
Thus, anti-transfer is a well-suited approach to exploit datasets provided with multiple labels, but the use of models pre-trained on different data is still effective and both can also be combined.

\paragraph{Related Work}
As mentioned in Section~\ref{previouswork}, when pre-training with an orthogonal label of the same dataset, AT is similar to Domain Adversarial Training (DAT) \cite{DBLP:journals/jmlr/GaninUAGLLML16} if we consider an orthogonal task class as a domain. 
As mentioned, AT has the practical advantage of only needing a pre-trained model rather than requiring labeled data from the source domain. 
This makes it possible to use models pre-trained by third parties, which can be beneficial in the case of models pre-trained on very large or private datasets. 
Even though in our test cases AT with models pre-trained on the same dataset provided the best improvement, models pre-trained on bigger and different datasets (which is not possible with DAT) still provided a good improvement over the baseline.

As mentioned in Section~\ref{previouswork}, the idea of anti-transfer is related to \textit{Speech-VGG} \cite{beckmann2019speech}, which applies a deep feature loss to encourage similarity of deep representations, instead of dissimilarity as in anti-transfer. 
The experiments by \cite{beckmann2019speech} are comparable with our inverse-Beta experiment in Section \ref{subsec:ablation}, where we show that encouraging similarity causes a drop in performance for orthogonal tasks. 
However, \cite{beckmann2019speech} obtain a performance improvement with their approach  applied to related target/pre-trained task combinations: word recognition vs. speech inpainting, language identification and speech/music classification.
This confirms that the selection of orthogonal tasks for anti-transfer is important.


As we introduce in Section~\ref{previouswork}, anti-transfer falls into the broad category of disentanglement. 
Our method does not directly enable pinpointing specific disentangled components in the data, e.g. as in source separation, but in effect it leads to separate deep representations for different tasks, as visualized in Figure \ref{fig:activations}. 
An advantage of anti-transfer is that it is a supervised training approach, which tends to be more efficient than adversarial or VAE methods. 

\paragraph{Limitations}
Limitations of anti-transfer apply to: resources, orthogonal tasks, models and data availability, pre-trained model accuracy. 

Anti-transfer needs additional memory and computation resources at training time. 
Invariance to simple transformations can sometimes be achieved with simpler models, e.g. \cite{DBLP:conf/icassp/GuizzoWL20,DBLP:conf/cvpr/WangKFYR19,DBLP:conf/mlsp/MarchandP16}, but complex tasks, like speaker recognition justifies in our view the increased resources used. 
Memory demands can have an impact in practice as GPU memory is often a bottleneck.  
Since earlier layers are similarly effective as later layers, but use less memory, using them can offer a better ratio of cost to performance gain. 
To make anti-transfer more practical on GPUs with limited memory, other ways of reducing memory demand can be investigated.

A practical limitation when using pre-trained networks, is that the target task network needs to have the same structure (up to the AT layer) as the orthogonal task network. 
This can be a limitation if the network structure is not well suited for the main task. 

Anti-transfer training is sensitive to the weight on the AT loss in our experiments, especially using the cosine similarity, although a value of $\beta = 1$ worked well in all our experiments. 
Still, some effort should be made to tune this hyper-parameter when using anti-transfer learning. 

A more conceptual limitation is the need for an orthogonal task.  However, identifying the orthogonal task is often straightforward, as the elements that cause model performance to decrease are known, e.g. speaker identity, text, emotion,  recording equipment, acoustical conditions. 
Finding or creating orthogonal task labels on the same or a similar dataset, or a model pre-trained on an orthogonal task, can be a limitation, depending on the application.

In addition to this, benefits of AT can only be expected if the pre-trained model is effective and even then there may be relevant representations that the pre-trained model has not learned. 
However, perfect avoidance of the representation learnt for the orthogonal task or perfect invariance to the orthogonal task is not required to improve performance and generalization, as our experiments have shown. 
The situation would be different for undesirable labels, where invariance to the orthogonal task in itself is an important target. 
Our measurement of this invariance has mainly been indirect through performance. 
Our visualization example was encouraging but to guarantee algorithmic fairness, more stringent measurements would be required. 

\paragraph{Applications}
Anti-transfer is in principle applicable in all situations where suitable datasets are available, in particular when invariance to a specific task is desired. 
Even though we implemented AT for VGG16,
it can be applied to other CNN designs. Also, it is directly applicable to feed-forward and to recurrent networks and it can be adapted to attention-based models.
We have only tested classification tasks, but there is nothing in general to prevent the application of this method to regression, or more complex tasks (e.g. automatic speech recognition) or other domains  (e.g. computer vision). 

As mentioned, AT can have applications in areas such as algorithmic fairness, where model outputs should be independent of sensitive variables, e.g. financial decisions should not depend on gender or ethnicity. 
The variable is not necessarily explicit in the input data, e.g. the gender of a person could be not mentioned in their financial data, but models could estimate it and use that estimate as the basis for a decision.
With more direct measurements of the degree of invariance to the sensitive variable,  AT could be suitable to improve algorithmic fairness. 

There are many pre-trained models available for many tasks, particularly in computer vision and natural language processing. 
These models can be used for anti-transfer in many tasks, with the limitation that the orthogonal task must be known to be independent of the target and the the network architecture must have  sufficient overlap, i.e. the structure of the networks must be the same from the input up to the the layer(s) used in anti-transfer.  



\section{Conclusions}
\label{conclusions}

In this study, we introduced anti-transfer learning for speech processing with neural networks, a novel method improving generalization by instilling invariance to an orthogonal task when training a network on a target task. 
When applying anti-transfer, we use a pre-trained network with the same structure as the target network. 
In training the target network we apply a deep feature loss that discourages similarity between convolutional layers in the pre-trained and target network to encourage the development of an internal representation that is independent of the orthogonal task. 
Our experiments with several classification tasks on speech and music audio in different configurations show  improved results for all tasks. 
We observe a robust improvement over the learning form scratch and over transfer learning by weight initialization. 


Our analysis provides evidence that anti-transfer achieves a degree of invariance to the orthogonal tasks, e.g. speaker identity, when the network is applied to the target task, e.g. speech emotion recognition. 
While there is a cost of pre-training and of the anti-transfer learning itself, the improved generalization may often be worth it. 
Readily available trained models remove the cost of pre-training and there may be further optimizations possible to address memory and computation costs. 

With the increasing availability of public datasets and pre-trained models chances grow that a suitable dataset or model can be found, but the selection of the orthogonal task needs careful consideration. 
Transfer learning is generally seen as a  straightforward way to improve the performance of deep neural networks by using additional data. 
Our results show that taking into account the nature of the pre-training tasks is important and that treating related and orthogonal tasks differently can boost generalization significantly. 

Applications can benefit from improved generalization in many domains where there are natural changes to a signal that are independent of the target task, such as room acoustics, ambient noise, degradation through transmission, etc., as in the tasks we addressed in our experiments. 
A potential application of anti-transfer is to avoid the use of specific signal properties in areas such as algorithmic fairness, where being invariant to gender or ethnicity is a socially important goal. 
This will need further work on measuring and controlling the level of invariance as well as a discussion of the specific goals. 

The positive results justify further investigation of this approach. 
An immediate research objective is to reduce the memory  requirements of anti-transfer learning.
Identifying more application areas and studying larger datasets in different domains will enable a better understanding of the performance an comparison to standard benchmarks. 
Further general goals for longer term research are a deeper understanding of how to measure invariance or achieve it across multiple tasks beyond dual anti-transfer and
it will be interesting to apply anti-transfer learning to
different Neural Network architectures, including non-convolutional ones. 

\section*{Acknowledgements}
\noindent This work is supported by City, University of London through a PhD studentship to Eric Guizzo.

\bibliography{bibliography_antitransfer}

\begin{thebibliography}{10}
\expandafter\ifx\csname url\endcsname\relax
  \def\url#1{\texttt{#1}}\fi
\expandafter\ifx\csname urlprefix\endcsname\relax\def\urlprefix{URL }\fi
\expandafter\ifx\csname href\endcsname\relax
  \def\href#1#2{#2} \def\path#1{#1}\fi

\bibitem{DBLP:conf/ismir/OordDS14}
A.~van~den Oord, S.~Dieleman, B.~Schrauwen,
  \href{http://www.terasoft.com.tw/conf/ismir2014/proceedings/T007\_118\_Paper.pdf}{Transfer
  learning by supervised pre-training for audio-based music classification},
  in: H.~Wang, Y.~Yang, J.~H. Lee (Eds.), Proceedings of the 15th International
  Society for Music Information Retrieval Conference, {ISMIR} 2014, Taipei,
  Taiwan, October 27-31, 2014, 2014, pp. 29--34.
\newline\urlprefix\url{http://www.terasoft.com.tw/conf/ismir2014/proceedings/T007\_118\_Paper.pdf}

\bibitem{DBLP:conf/naacl/BansalKLLG19}
S.~Bansal, H.~Kamper, K.~Livescu, A.~Lopez, S.~Goldwater,
  \href{https://doi.org/10.18653/v1/n19-1006}{Pre-training on high-resource
  speech recognition improves low-resource speech-to-text translation}, in:
  J.~Burstein, C.~Doran, T.~Solorio (Eds.), Proceedings of the 2019 Conference
  of the North American Chapter of the Association for Computational
  Linguistics: Human Language Technologies, {NAACL-HLT} 2019, Minneapolis, MN,
  USA, June 2-7, 2019, Volume 1 (Long and Short Papers), Association for
  Computational Linguistics, 2019, pp. 58--68.
\newblock \href {http://dx.doi.org/10.18653/v1/n19-1006}
  {\path{doi:10.18653/v1/n19-1006}}.
\newline\urlprefix\url{https://doi.org/10.18653/v1/n19-1006}

\bibitem{DBLP:conf/ismir/DielemanBS11}
S.~Dieleman, P.~Brakel, B.~Schrauwen,
  \href{http://ismir2011.ismir.net/papers/PS6-3.pdf}{Audio-based music
  classification with a pretrained convolutional network}, in: A.~Klapuri,
  C.~Leider (Eds.), Proceedings of the 12th International Society for Music
  Information Retrieval Conference, {ISMIR} 2011, Miami, Florida, USA, October
  24-28, 2011, University of Miami, 2011, pp. 669--674.
\newline\urlprefix\url{http://ismir2011.ismir.net/papers/PS6-3.pdf}

\bibitem{DBLP:journals/corr/abs-1909-10924}
P.~Wang, L.~Wei, Y.~Cao, J.~Xie, Y.~Cao, Z.~Nie,
  \href{http://arxiv.org/abs/1909.10924}{Understanding semantics from speech
  through pre-training}, CoRR abs/1909.10924.
\newblock \href {http://arxiv.org/abs/1909.10924} {\path{arXiv:1909.10924}}.
\newline\urlprefix\url{http://arxiv.org/abs/1909.10924}

\bibitem{DBLP:conf/interspeech/KitzaSN18}
M.~Kitza, R.~Schl{\"{u}}ter, H.~Ney,
  \href{https://doi.org/10.21437/Interspeech.2018-2022}{Comparison of
  blstm-layer-specific affine transformations for speaker adaptation}, in:
  B.~Yegnanarayana (Ed.), Interspeech 2018, 19th Annual Conference of the
  International Speech Communication Association, Hyderabad, India, 2-6
  September 2018, {ISCA}, 2018, pp. 877--881.
\newblock \href {http://dx.doi.org/10.21437/Interspeech.2018-2022}
  {\path{doi:10.21437/Interspeech.2018-2022}}.
\newline\urlprefix\url{https://doi.org/10.21437/Interspeech.2018-2022}

\bibitem{DBLP:conf/interspeech/MildeB18}
B.~Milde, C.~Biemann,
  \href{https://doi.org/10.21437/Interspeech.2018-2194}{Unspeech: Unsupervised
  speech context embeddings}, in: B.~Yegnanarayana (Ed.), Interspeech 2018,
  19th Annual Conference of the International Speech Communication Association,
  Hyderabad, India, 2-6 September 2018, {ISCA}, 2018, pp. 2693--2697.
\newblock \href {http://dx.doi.org/10.21437/Interspeech.2018-2194}
  {\path{doi:10.21437/Interspeech.2018-2194}}.
\newline\urlprefix\url{https://doi.org/10.21437/Interspeech.2018-2194}

\bibitem{DBLP:conf/icira/LiuXLH19}
Z.~Liu, P.~Xiao, D.~Li, M.~Hao,
  \href{https://doi.org/10.1007/978-3-030-27535-8\_43}{Speaker-independent
  speech emotion recognition based on {CNN-BLSTM} and multiple svms}, in:
  H.~Yu, J.~Liu, L.~Liu, Z.~Ju, Y.~Liu, D.~Zhou (Eds.), Intelligent Robotics
  and Applications - 12th International Conference, {ICIRA} 2019, Shenyang,
  China, August 8-11, 2019, Proceedings, Part {III}, Vol. 11742 of Lecture
  Notes in Computer Science, Springer, 2019, pp. 481--491.
\newblock \href {http://dx.doi.org/10.1007/978-3-030-27535-8\_43}
  {\path{doi:10.1007/978-3-030-27535-8\_43}}.
\newline\urlprefix\url{https://doi.org/10.1007/978-3-030-27535-8\_43}

\bibitem{DBLP:conf/asru/JalalMH19}
M.~A. Jalal, R.~K. Moore, T.~Hain,
  \href{https://doi.org/10.1109/ASRU46091.2019.9004037}{Spatio-temporal context
  modelling for speech emotion classification}, in: {IEEE} Automatic Speech
  Recognition and Understanding Workshop, {ASRU} 2019, Singapore, December
  14-18, 2019, {IEEE}, 2019, pp. 853--859.
\newblock \href {http://dx.doi.org/10.1109/ASRU46091.2019.9004037}
  {\path{doi:10.1109/ASRU46091.2019.9004037}}.
\newline\urlprefix\url{https://doi.org/10.1109/ASRU46091.2019.9004037}

\bibitem{DBLP:journals/amcs/RybkaJ13}
J.~Rybka, A.~Janicki, \href{https://doi.org/10.2478/amcs-2013-0060}{Comparison
  of speaker dependent and speaker independent emotion recognition}, Int. J.
  Appl. Math. Comput. Sci. 23~(4) (2013) 797--808.
\newblock \href {http://dx.doi.org/10.2478/amcs-2013-0060}
  {\path{doi:10.2478/amcs-2013-0060}}.
\newline\urlprefix\url{https://doi.org/10.2478/amcs-2013-0060}

\bibitem{speakerdependent2013}
M.~Bhaykar, J.~Yadav, K.~Rao, Speaker dependent, speaker independent and cross
  language emotion recognition from speech using gmm and hmm, 2013, pp. 1--5.
\newblock \href {http://dx.doi.org/10.1109/NCC.2013.6487998}
  {\path{doi:10.1109/NCC.2013.6487998}}.

\bibitem{DBLP:conf/icann/TanSKZYL18}
C.~Tan, F.~Sun, T.~Kong, W.~Zhang, C.~Yang, C.~Liu,
  \href{https://doi.org/10.1007/978-3-030-01424-7\_27}{A survey on deep
  transfer learning}, in: V.~Kurkov{\'{a}}, Y.~Manolopoulos, B.~Hammer, L.~S.
  Iliadis, I.~Maglogiannis (Eds.), Artificial Neural Networks and Machine
  Learning - {ICANN} 2018 - 27th International Conference on Artificial Neural
  Networks, Rhodes, Greece, October 4-7, 2018, Proceedings, Part {III}, Vol.
  11141 of Lecture Notes in Computer Science, Springer, 2018, pp. 270--279.
\newblock \href {http://dx.doi.org/10.1007/978-3-030-01424-7\_27}
  {\path{doi:10.1007/978-3-030-01424-7\_27}}.
\newline\urlprefix\url{https://doi.org/10.1007/978-3-030-01424-7\_27}

\bibitem{caruana1995learning}
R.~Caruana, Learning many related tasks at the same time with backpropagation,
  in: Advances in neural information processing systems, 1995, pp. 657--664.

\bibitem{bengio2012deep}
Y.~Bengio, \href{http://proceedings.mlr.press/v27/bengio12a.html}{Deep learning
  of representations for unsupervised and transfer learning}, in: I.~Guyon,
  G.~Dror, V.~Lemaire, G.~W. Taylor, D.~L. Silver (Eds.), Unsupervised and
  Transfer Learning - Workshop held at {ICML} 2011, Bellevue, Washington, USA,
  July 2, 2011, Vol.~27 of {JMLR} Proceedings, JMLR.org, 2012, pp. 17--36.
\newline\urlprefix\url{http://proceedings.mlr.press/v27/bengio12a.html}

\bibitem{41530}
P.~Hamel, M.~E.~P. Davies, K.~Yoshii, M.~Goto,
  \href{http://www.ppgia.pucpr.br/ismir2013/wp-content/uploads/2013/09/76\_Paper.pdf}{Transfer
  learning in {MIR}: Sharing learned latent representations for music audio
  classification and similarity}, in: A.~de~Souza Britto~Jr., F.~Gouyon,
  S.~Dixon (Eds.), Proceedings of the 14th International Society for Music
  Information Retrieval Conference, {ISMIR} 2013, Curitiba, Brazil, November
  4-8, 2013, 2013, pp. 9--14.
\newline\urlprefix\url{http://www.ppgia.pucpr.br/ismir2013/wp-content/uploads/2013/09/76\_Paper.pdf}

\bibitem{shin2016deep}
H.~Shin, H.~R. Roth, M.~Gao, L.~Lu, Z.~Xu, I.~Nogues, J.~Yao, D.~J. Mollura,
  R.~M. Summers, \href{https://doi.org/10.1109/TMI.2016.2528162}{Deep
  convolutional neural networks for computer-aided detection: {CNN}
  architectures, dataset characteristics and transfer learning}, {IEEE} Trans.
  Medical Imaging 35~(5) (2016) 1285--1298.
\newblock \href {http://dx.doi.org/10.1109/TMI.2016.2528162}
  {\path{doi:10.1109/TMI.2016.2528162}}.
\newline\urlprefix\url{https://doi.org/10.1109/TMI.2016.2528162}

\bibitem{tan2018survey}
C.~Tan, F.~Sun, T.~Kong, W.~Zhang, C.~Yang, C.~Liu,
  \href{https://doi.org/10.1007/978-3-030-01424-7\_27}{A survey on deep
  transfer learning}, in: V.~Kurkov{\'{a}}, Y.~Manolopoulos, B.~Hammer, L.~S.
  Iliadis, I.~Maglogiannis (Eds.), Artificial Neural Networks and Machine
  Learning - {ICANN} 2018 - 27th International Conference on Artificial Neural
  Networks, Rhodes, Greece, October 4-7, 2018, Proceedings, Part {III}, Vol.
  11141 of Lecture Notes in Computer Science, Springer, 2018, pp. 270--279.
\newblock \href {http://dx.doi.org/10.1007/978-3-030-01424-7\_27}
  {\path{doi:10.1007/978-3-030-01424-7\_27}}.
\newline\urlprefix\url{https://doi.org/10.1007/978-3-030-01424-7\_27}

\bibitem{DBLP:conf/icdar/StuderAPGK0LI19}
L.~Studer, M.~Alberti, V.~Pondenkandath, P.~Goktepe, T.~Kolonko, A.~Fischer,
  M.~Liwicki, R.~Ingold, \href{https://doi.org/10.1109/ICDAR.2019.00120}{A
  comprehensive study of imagenet pre-training for historical document image
  analysis}, in: 2019 International Conference on Document Analysis and
  Recognition, {ICDAR} 2019, Sydney, Australia, September 20-25, 2019, {IEEE},
  2019, pp. 720--725.
\newblock \href {http://dx.doi.org/10.1109/ICDAR.2019.00120}
  {\path{doi:10.1109/ICDAR.2019.00120}}.
\newline\urlprefix\url{https://doi.org/10.1109/ICDAR.2019.00120}

\bibitem{DBLP:conf/eccv/XieR18}
Y.~Xie, D.~Richmond,
  \href{https://doi.org/10.1007/978-3-030-11024-6\_37}{Pre-training on
  grayscale imagenet improves medical image classification}, in:
  L.~Leal{-}Taix{\'{e}}, S.~Roth (Eds.), Computer Vision - {ECCV} 2018
  Workshops - Munich, Germany, September 8-14, 2018, Proceedings, Part {VI},
  Vol. 11134 of Lecture Notes in Computer Science, Springer, 2018, pp.
  476--484.
\newblock \href {http://dx.doi.org/10.1007/978-3-030-11024-6\_37}
  {\path{doi:10.1007/978-3-030-11024-6\_37}}.
\newline\urlprefix\url{https://doi.org/10.1007/978-3-030-11024-6\_37}

\bibitem{DBLP:journals/amm/HanJMX18}
Y.~Han, T.~Jiang, Y.~Ma, C.~Xu,
  \href{https://doi.org/10.1155/2018/3138278}{Pretraining convolutional neural
  networks for image-based vehicle classification}, Adv. Multim. 2018 (2018)
  3138278:1--3138278:10.
\newblock \href {http://dx.doi.org/10.1155/2018/3138278}
  {\path{doi:10.1155/2018/3138278}}.
\newline\urlprefix\url{https://doi.org/10.1155/2018/3138278}

\bibitem{DBLP:journals/corr/abs-2003-08271}
X.~Qiu, T.~Sun, Y.~Xu, Y.~Shao, N.~Dai, X.~Huang,
  \href{https://arxiv.org/abs/2003.08271}{Pre-trained models for natural
  language processing: {A} survey}, CoRR abs/2003.08271.
\newblock \href {http://arxiv.org/abs/2003.08271} {\path{arXiv:2003.08271}}.
\newline\urlprefix\url{https://arxiv.org/abs/2003.08271}

\bibitem{Pratt-1992-Discriminability-Based}
L.~Y. Pratt,
  \href{http://papers.nips.cc/paper\/641-discriminability-based-transfer-between-neural-networks}{Discriminability-based
  transfer between neural networks}, in: S.~J. Hanson, J.~D. Cowan, C.~L. Giles
  (Eds.), Advances in Neural Information Processing Systems 5, {[NIPS}
  Conference, Denver, Colorado, USA, November 30 - December 3, 1992], Morgan
  Kaufmann, 1992, pp. 204--211.
\newline\urlprefix\url{http://papers.nips.cc/paper\/641-discriminability-based-transfer-between-neural-networks}

\bibitem{Gutstein-et-al-2007-Knowledge-Transfer}
S.~Gutstein, O.~Fuentes, E.~Freudenthal,
  \href{http://www.aaai.org/Library/FLAIRS/2007/flairs07-020.php}{Knowledge
  transfer in deep convolutional neural nets}, in: D.~Wilson, G.~Sutcliffe
  (Eds.), Proceedings of the Twentieth International Florida Artificial
  Intelligence Research Society Conference, May 7-9, 2007, Key West, Florida,
  {USA}, {AAAI} Press, 2007, pp. 104--109.
\newline\urlprefix\url{http://www.aaai.org/Library/FLAIRS/2007/flairs07-020.php}

\bibitem{gatys2016image}
L.~A. Gatys, A.~S. Ecker, M.~Bethge,
  \href{https://doi.org/10.1109/CVPR.2016.265}{Image style transfer using
  convolutional neural networks}, in: 2016 {IEEE} Conference on Computer Vision
  and Pattern Recognition, {CVPR} 2016, Las Vegas, NV, USA, June 27-30, 2016,
  {IEEE} Computer Society, 2016, pp. 2414--2423.
\newblock \href {http://dx.doi.org/10.1109/CVPR.2016.265}
  {\path{doi:10.1109/CVPR.2016.265}}.
\newline\urlprefix\url{https://doi.org/10.1109/CVPR.2016.265}

\bibitem{yosinski2014transferable}
J.~Yosinski, J.~Clune, Y.~Bengio, H.~Lipson,
  \href{http://papers.nips.cc/paper/5347-how-transferable-are-features-in-deep-neural-networks}{How
  transferable are features in deep neural networks?}, in: Z.~Ghahramani,
  M.~Welling, C.~Cortes, N.~D. Lawrence, K.~Q. Weinberger (Eds.), Advances in
  Neural Information Processing Systems 27: Annual Conference on Neural
  Information Processing Systems 2014, December 8-13 2014, Montreal, Quebec,
  Canada, 2014, pp. 3320--3328.
\newline\urlprefix\url{http://papers.nips.cc/paper/5347-how-transferable-are-features-in-deep-neural-networks}

\bibitem{ulyanov2016texture}
D.~Ulyanov, V.~Lebedev, A.~Vedaldi, V.~S. Lempitsky,
  \href{http://proceedings.mlr.press/v48/ulyanov16.html}{Texture networks:
  Feed-forward synthesis of textures and stylized images}, in: M.~Balcan, K.~Q.
  Weinberger (Eds.), Proceedings of the 33nd International Conference on
  Machine Learning, {ICML} 2016, New York City, NY, USA, June 19-24, 2016,
  Vol.~48 of {JMLR} Workshop and Conference Proceedings, JMLR.org, 2016, pp.
  1349--1357.
\newline\urlprefix\url{http://proceedings.mlr.press/v48/ulyanov16.html}

\bibitem{dumoulin2016learned}
V.~Dumoulin, J.~Shlens, M.~Kudlur,
  \href{https://openreview.net/forum?id=BJO-BuT1g}{A learned representation for
  artistic style}, in: 5th International Conference on Learning
  Representations, {ICLR} 2017, Toulon, France, April 24-26, 2017, Conference
  Track Proceedings, OpenReview.net, 2017.
\newline\urlprefix\url{https://openreview.net/forum?id=BJO-BuT1g}

\bibitem{pasini2019melgan}
M.~Pasini, \href{http://arxiv.org/abs/1910.03713}{Melgan-vc: Voice conversion
  and audio style transfer on arbitrarily long samples using spectrograms},
  CoRR abs/1910.03713.
\newblock \href {http://arxiv.org/abs/1910.03713} {\path{arXiv:1910.03713}}.
\newline\urlprefix\url{http://arxiv.org/abs/1910.03713}

\bibitem{gatys2015texture}
L.~A. Gatys, A.~S. Ecker, M.~Bethge,
  \href{http://papers.nips.cc/paper/5633-texture-synthesis-using-convolutional-neural-networks}{Texture
  synthesis using convolutional neural networks}, in: C.~Cortes, N.~D.
  Lawrence, D.~D. Lee, M.~Sugiyama, R.~Garnett (Eds.), Advances in Neural
  Information Processing Systems 28: Annual Conference on Neural Information
  Processing Systems 2015, December 7-12, 2015, Montreal, Quebec, Canada, 2015,
  pp. 262--270.
\newline\urlprefix\url{http://papers.nips.cc/paper/5633-texture-synthesis-using-convolutional-neural-networks}

\bibitem{johnson2016perceptual}
J.~Johnson, A.~Alahi, L.~Fei{-}Fei,
  \href{https://doi.org/10.1007/978-3-319-46475-6\_43}{Perceptual losses for
  real-time style transfer and super-resolution}, in: B.~Leibe, J.~Matas,
  N.~Sebe, M.~Welling (Eds.), Computer Vision - {ECCV} 2016 - 14th European
  Conference, Amsterdam, The Netherlands, October 11-14, 2016, Proceedings,
  Part {II}, Vol. 9906 of Lecture Notes in Computer Science, Springer, 2016,
  pp. 694--711.
\newblock \href {http://dx.doi.org/10.1007/978-3-319-46475-6\_43}
  {\path{doi:10.1007/978-3-319-46475-6\_43}}.
\newline\urlprefix\url{https://doi.org/10.1007/978-3-319-46475-6\_43}

\bibitem{chen2017photographic}
Q.~Chen, V.~Koltun, \href{https://doi.org/10.1109/ICCV.2017.168}{Photographic
  image synthesis with cascaded refinement networks}, in: {IEEE} International
  Conference on Computer Vision, {ICCV} 2017, Venice, Italy, October 22-29,
  2017, {IEEE} Computer Society, 2017, pp. 1520--1529.
\newblock \href {http://dx.doi.org/10.1109/ICCV.2017.168}
  {\path{doi:10.1109/ICCV.2017.168}}.
\newline\urlprefix\url{https://doi.org/10.1109/ICCV.2017.168}

\bibitem{dosovitskiy2016generating}
A.~Dosovitskiy, T.~Brox,
  \href{http://papers.nips.cc/paper/6158-generating-images-with-perceptual-similarity-metrics-based-on-deep-networks}{Generating
  images with perceptual similarity metrics based on deep networks}, in: D.~D.
  Lee, M.~Sugiyama, U.~von Luxburg, I.~Guyon, R.~Garnett (Eds.), Advances in
  Neural Information Processing Systems 29: Annual Conference on Neural
  Information Processing Systems 2016, December 5-10, 2016, Barcelona, Spain,
  2016, pp. 658--666.
\newline\urlprefix\url{http://papers.nips.cc/paper/6158-generating-images-with-perceptual-similarity-metrics-based-on-deep-networks}

\bibitem{zhang2018unreasonable}
R.~Zhang, P.~Isola, A.~A. Efros, E.~Shechtman, O.~Wang,
  \href{http://openaccess.thecvf.com/content\_cvpr\_2018/html/Zhang\_The\_Unreasonable\_Effectiveness\_CVPR\_2018\_paper.html}{The
  unreasonable effectiveness of deep features as a perceptual metric}, in: 2018
  {IEEE} Conference on Computer Vision and Pattern Recognition, {CVPR} 2018,
  Salt Lake City, UT, USA, June 18-22, 2018, {IEEE} Computer Society, 2018, pp.
  586--595.
\newblock \href {http://dx.doi.org/10.1109/CVPR.2018.00068}
  {\path{doi:10.1109/CVPR.2018.00068}}.
\newline\urlprefix\url{http://openaccess.thecvf.com/content\_cvpr\_2018/html/Zhang\_The\_Unreasonable\_Effectiveness\_CVPR\_2018\_paper.html}

\bibitem{doersch2015unsupervised}
C.~Doersch, A.~Gupta, A.~A. Efros,
  \href{https://doi.org/10.1109/ICCV.2015.167}{Unsupervised visual
  representation learning by context prediction}, in: 2015 {IEEE} International
  Conference on Computer Vision, {ICCV} 2015, Santiago, Chile, December 7-13,
  2015, {IEEE} Computer Society, 2015, pp. 1422--1430.
\newblock \href {http://dx.doi.org/10.1109/ICCV.2015.167}
  {\path{doi:10.1109/ICCV.2015.167}}.
\newline\urlprefix\url{https://doi.org/10.1109/ICCV.2015.167}

\bibitem{beckmann2019speech}
P.~Beckmann, M.~Kegler, H.~Saltini, M.~Cernak,
  \href{http://arxiv.org/abs/1910.09909}{Speech-{VGG}: {A} deep feature
  extractor for speech processing}, CoRR abs/1910.09909.
\newblock \href {http://arxiv.org/abs/1910.09909} {\path{arXiv:1910.09909}}.
\newline\urlprefix\url{http://arxiv.org/abs/1910.09909}

\bibitem{sahai2019spectrogram}
A.~Sahai, R.~Weber, B.~McWilliams,
  \href{https://doi.org/10.23919/EUSIPCO.2019.8903019}{Spectrogram feature
  losses for music source separation}, in: 27th European Signal Processing
  Conference, {EUSIPCO} 2019, {A} Coru{\~{n}}a, Spain, September 2-6, 2019,
  {IEEE}, 2019, pp. 1--5.
\newblock \href {http://dx.doi.org/10.23919/EUSIPCO.2019.8903019}
  {\path{doi:10.23919/EUSIPCO.2019.8903019}}.
\newline\urlprefix\url{https://doi.org/10.23919/EUSIPCO.2019.8903019}

\bibitem{kegler2019deep}
M.~Kegler, P.~Beckmann, M.~Cernak, \href{http://arxiv.org/abs/1910.09058}{Deep
  speech inpainting of time-frequency masks}, CoRR abs/1910.09058.
\newblock \href {http://arxiv.org/abs/1910.09058} {\path{arXiv:1910.09058}}.
\newline\urlprefix\url{http://arxiv.org/abs/1910.09058}

\bibitem{yao2004evolving}
X.~Yao, Y.~Liu,
  \href{http://content.iospress.com/articles/international-journal-of-hybrid-intelligent-systems/his004}{Evolving
  neural network ensembles by minimization of mutual information}, Int. J.
  Hybrid Intell. Syst. 1~(1) (2004) 12--21.
\newline\urlprefix\url{http://content.iospress.com/articles/international-journal-of-hybrid-intelligent-systems/his004}

\bibitem{liu2018learning}
W.~Liu, R.~Lin, Z.~Liu, L.~Liu, Z.~Yu, B.~Dai, L.~Song,
  \href{http://papers.nips.cc/paper/7860-learning-towards-minimum-hyperspherical-energy}{Learning
  towards minimum hyperspherical energy}, in: S.~Bengio, H.~M. Wallach,
  H.~Larochelle, K.~Grauman, N.~Cesa{-}Bianchi, R.~Garnett (Eds.), Advances in
  Neural Information Processing Systems 31: Annual Conference on Neural
  Information Processing Systems 2018, NeurIPS 2018, 3-8 December 2018,
  Montr{\'{e}}al, Canada, 2018, pp. 6225--6236.
\newline\urlprefix\url{http://papers.nips.cc/paper/7860-learning-towards-minimum-hyperspherical-energy}

\bibitem{perez2020improving}
J.~Perez{-}Lapillo, O.~Galkin, T.~Weyde,
  \href{https://doi.org/10.1109/ICASSP40776.2020.9053424}{Improving singing
  voice separation with the wave-u-net using minimum hyperspherical energy},
  in: 2020 {IEEE} International Conference on Acoustics, Speech and Signal
  Processing, {ICASSP} 2020, Barcelona, Spain, May 4-8, 2020, {IEEE}, 2020, pp.
  3272--3276.
\newblock \href {http://dx.doi.org/10.1109/ICASSP40776.2020.9053424}
  {\path{doi:10.1109/ICASSP40776.2020.9053424}}.
\newline\urlprefix\url{https://doi.org/10.1109/ICASSP40776.2020.9053424}

\bibitem{wang2019learning}
K.~Wang, J.~Liu, J.~Wang, \href{https://doi.org/10.1155/2019/9414539}{Learning
  domain-independent deep representations by mutual information minimization},
  Comput. Intell. Neurosci. 2019 (2019) 9414539:1--9414539:14.
\newblock \href {http://dx.doi.org/10.1155/2019/9414539}
  {\path{doi:10.1155/2019/9414539}}.
\newline\urlprefix\url{https://doi.org/10.1155/2019/9414539}

\bibitem{DBLP:journals/jmlr/GaninUAGLLML16}
Y.~Ganin, E.~Ustinova, H.~Ajakan, P.~Germain, H.~Larochelle, F.~Laviolette,
  M.~Marchand, V.~S. Lempitsky,
  \href{http://jmlr.org/papers/v17/15-239.html}{Domain-adversarial training of
  neural networks}, J. Mach. Learn. Res. 17 (2016) 59:1--59:35.
\newline\urlprefix\url{http://jmlr.org/papers/v17/15-239.html}

\bibitem{DBLP:conf/iccv/MotiianPAD17}
S.~Motiian, M.~Piccirilli, D.~A. Adjeroh, G.~Doretto,
  \href{https://doi.org/10.1109/ICCV.2017.609}{Unified deep supervised domain
  adaptation and generalization}, in: {IEEE} International Conference on
  Computer Vision, {ICCV} 2017, Venice, Italy, October 22-29, 2017, {IEEE}
  Computer Society, 2017, pp. 5716--5726.
\newblock \href {http://dx.doi.org/10.1109/ICCV.2017.609}
  {\path{doi:10.1109/ICCV.2017.609}}.
\newline\urlprefix\url{https://doi.org/10.1109/ICCV.2017.609}

\bibitem{DBLP:conf/iclr/TzengHSD17}
E.~Tzeng, J.~Hoffman, K.~Saenko, T.~Darrell,
  \href{https://openreview.net/forum?id=B1Vjl1Stl}{Adversarial discriminative
  domain adaptation (workshop extended abstract)}, in: 5th International
  Conference on Learning Representations, {ICLR} 2017, Toulon, France, April
  24-26, 2017, Workshop Track Proceedings, OpenReview.net, 2017.
\newline\urlprefix\url{https://openreview.net/forum?id=B1Vjl1Stl}

\bibitem{DBLP:conf/waspaa/DrossosMV19}
K.~Drossos, P.~Magron, T.~Virtanen,
  \href{https://doi.org/10.1109/WASPAA.2019.8937231}{Unsupervised adversarial
  domain adaptation based on the wasserstein distance for acoustic scene
  classification}, in: 2019 {IEEE} Workshop on Applications of Signal
  Processing to Audio and Acoustics, {WASPAA} 2019, New Paltz, NY, USA, October
  20-23, 2019, {IEEE}, 2019, pp. 259--263.
\newblock \href {http://dx.doi.org/10.1109/WASPAA.2019.8937231}
  {\path{doi:10.1109/WASPAA.2019.8937231}}.
\newline\urlprefix\url{https://doi.org/10.1109/WASPAA.2019.8937231}

\bibitem{DBLP:journals/corr/abs-1811-03271}
Y.~Hung, Y.~Chen, Y.~Yang, \href{http://arxiv.org/abs/1811.03271}{Learning
  disentangled representations for timber and pitch in music audio}, CoRR
  abs/1811.03271.
\newblock \href {http://arxiv.org/abs/1811.03271} {\path{arXiv:1811.03271}}.
\newline\urlprefix\url{http://arxiv.org/abs/1811.03271}

\bibitem{DBLP:conf/interspeech/ChouYLL18}
J.~Chou, C.~Yeh, H.~Lee, L.~Lee,
  \href{https://doi.org/10.21437/Interspeech.2018-1830}{Multi-target voice
  conversion without parallel data by adversarially learning disentangled audio
  representations}, in: B.~Yegnanarayana (Ed.), Interspeech 2018, 19th Annual
  Conference of the International Speech Communication Association, Hyderabad,
  India, 2-6 September 2018, {ISCA}, 2018, pp. 501--505.
\newblock \href {http://dx.doi.org/10.21437/Interspeech.2018-1830}
  {\path{doi:10.21437/Interspeech.2018-1830}}.
\newline\urlprefix\url{https://doi.org/10.21437/Interspeech.2018-1830}

\bibitem{DBLP:conf/icassp/NagraniCAZ20}
A.~Nagrani, J.~S. Chung, S.~Albanie, A.~Zisserman,
  \href{https://doi.org/10.1109/ICASSP40776.2020.9054057}{Disentangled speech
  embeddings using cross-modal self-supervision}, in: 2020 {IEEE} International
  Conference on Acoustics, Speech and Signal Processing, {ICASSP} 2020,
  Barcelona, Spain, May 4-8, 2020, {IEEE}, 2020, pp. 6829--6833.
\newblock \href {http://dx.doi.org/10.1109/ICASSP40776.2020.9054057}
  {\path{doi:10.1109/ICASSP40776.2020.9054057}}.
\newline\urlprefix\url{https://doi.org/10.1109/ICASSP40776.2020.9054057}

\bibitem{DBLP:conf/icassp/LeeBSJN20}
J.~Lee, N.~J. Bryan, J.~Salamon, Z.~Jin, J.~Nam,
  \href{https://doi.org/10.1109/ICASSP40776.2020.9053442}{Disentangled
  multidimensional metric learning for music similarity}, in: 2020 {IEEE}
  International Conference on Acoustics, Speech and Signal Processing, {ICASSP}
  2020, Barcelona, Spain, May 4-8, 2020, {IEEE}, 2020, pp. 6--10.
\newblock \href {http://dx.doi.org/10.1109/ICASSP40776.2020.9053442}
  {\path{doi:10.1109/ICASSP40776.2020.9053442}}.
\newline\urlprefix\url{https://doi.org/10.1109/ICASSP40776.2020.9053442}

\bibitem{DBLP:journals/corr/abs-1904-04772}
J.~Oldfield, Y.~Panagakis, M.~A. Nicolaou,
  \href{http://arxiv.org/abs/1904.04772}{Adversarial learning of disentangled
  and generalizable representations for visual attributes}, CoRR
  abs/1904.04772.
\newblock \href {http://arxiv.org/abs/1904.04772} {\path{arXiv:1904.04772}}.
\newline\urlprefix\url{http://arxiv.org/abs/1904.04772}

\bibitem{DBLP:conf/icml/LiM18a}
Y.~Li, S.~Mandt,
  \href{http://proceedings.mlr.press/v80/yingzhen18a.html}{Disentangled
  sequential autoencoder}, in: J.~G. Dy, A.~Krause (Eds.), Proceedings of the
  35th International Conference on Machine Learning, {ICML} 2018,
  Stockholmsm{\"{a}}ssan, Stockholm, Sweden, July 10-15, 2018, Vol.~80 of
  Proceedings of Machine Learning Research, {PMLR}, 2018, pp. 5656--5665.
\newline\urlprefix\url{http://proceedings.mlr.press/v80/yingzhen18a.html}

\bibitem{DBLP:journals/corr/abs-1805-11264}
W.~Hsu, J.~R. Glass, \href{http://arxiv.org/abs/1805.11264}{Disentangling by
  partitioning: {A} representation learning framework for multimodal sensory
  data}, CoRR abs/1805.11264.
\newblock \href {http://arxiv.org/abs/1805.11264} {\path{arXiv:1805.11264}}.
\newline\urlprefix\url{http://arxiv.org/abs/1805.11264}

\bibitem{DBLP:conf/ismir/JanssonHMBKW17}
A.~Jansson, E.~J. Humphrey, N.~Montecchio, R.~M. Bittner, A.~Kumar, T.~Weyde,
  \href{https://ismir2017.smcnus.org/wp-content/uploads/2017/10/171\_Paper.pdf}{Singing
  voice separation with deep u-net convolutional networks}, in: S.~J.
  Cunningham, Z.~Duan, X.~Hu, D.~Turnbull (Eds.), Proceedings of the 18th
  International Society for Music Information Retrieval Conference, {ISMIR}
  2017, Suzhou, China, October 23-27, 2017, 2017, pp. 745--751.
\newline\urlprefix\url{https://ismir2017.smcnus.org/wp-content/uploads/2017/10/171\_Paper.pdf}

\bibitem{DBLP:conf/interspeech/NeumannV17}
M.~Neumann, N.~T. Vu,
  \href{http://www.isca-speech.org/archive/Interspeech\_2017/abstracts/0917.html}{Attentive
  convolutional neural network based speech emotion recognition: {A} study on
  the impact of input features, signal length, and acted speech}, in:
  F.~Lacerda (Ed.), Interspeech 2017, 18th Annual Conference of the
  International Speech Communication Association, Stockholm, Sweden, August
  20-24, 2017, {ISCA}, 2017, pp. 1263--1267.
\newline\urlprefix\url{http://www.isca-speech.org/archive/Interspeech\_2017/abstracts/0917.html}

\bibitem{DBLP:journals/spl/SalamonB17}
J.~Salamon, J.~P. Bello, \href{https://doi.org/10.1109/LSP.2017.2657381}{Deep
  convolutional neural networks and data augmentation for environmental sound
  classification}, {IEEE} Signal Process. Lett. 24~(3) (2017) 279--283.
\newblock \href {http://dx.doi.org/10.1109/LSP.2017.2657381}
  {\path{doi:10.1109/LSP.2017.2657381}}.
\newline\urlprefix\url{https://doi.org/10.1109/LSP.2017.2657381}

\bibitem{simonyan2014very}
K.~Simonyan, A.~Zisserman, \href{http://arxiv.org/abs/1409.1556}{Very deep
  convolutional networks for large-scale image recognition}, in: Y.~Bengio,
  Y.~LeCun (Eds.), 3rd International Conference on Learning Representations,
  {ICLR} 2015, San Diego, CA, USA, May 7-9, 2015, Conference Track Proceedings,
  2015.
\newline\urlprefix\url{http://arxiv.org/abs/1409.1556}

\bibitem{romani_picas_oriol_2017_820937}
O.~Romani~Picas, H.~Parra~Rodriguez, D.~Dabiri, X.~Serra,
  \href{https://doi.org/10.5281/zenodo.820937}{Good-sounds dataset} (Jun.
  2017).
\newblock \href {http://dx.doi.org/10.5281/zenodo.820937}
  {\path{doi:10.5281/zenodo.820937}}.
\newline\urlprefix\url{https://doi.org/10.5281/zenodo.820937}

\bibitem{warden2018speech}
P.~Warden, \href{http://arxiv.org/abs/1804.03209}{Speech commands: {A} dataset
  for limited-vocabulary speech recognition}, CoRR abs/1804.03209.
\newblock \href {http://arxiv.org/abs/1804.03209} {\path{arXiv:1804.03209}}.
\newline\urlprefix\url{http://arxiv.org/abs/1804.03209}

\bibitem{reddy2019scalable}
C.~K.~A. Reddy, E.~Beyrami, J.~Pool, R.~Cutler, S.~Srinivasan, J.~Gehrke,
  \href{https://doi.org/10.21437/Interspeech.2019-3087}{A scalable noisy speech
  dataset and online subjective test framework}, in: G.~Kubin, Z.~Kacic (Eds.),
  Interspeech 2019, 20th Annual Conference of the International Speech
  Communication Association, Graz, Austria, 15-19 September 2019, {ISCA}, 2019,
  pp. 1816--1820.
\newblock \href {http://dx.doi.org/10.21437/Interspeech.2019-3087}
  {\path{doi:10.21437/Interspeech.2019-3087}}.
\newline\urlprefix\url{https://doi.org/10.21437/Interspeech.2019-3087}

\bibitem{Panayotov2015LibrispeechAA}
V.~Panayotov, G.~Chen, D.~Povey, S.~Khudanpur,
  \href{https://doi.org/10.1109/ICASSP.2015.7178964}{Librispeech: An {ASR}
  corpus based on public domain audio books}, in: 2015 {IEEE} International
  Conference on Acoustics, Speech and Signal Processing, {ICASSP} 2015, South
  Brisbane, Queensland, Australia, April 19-24, 2015, {IEEE}, 2015, pp.
  5206--5210.
\newblock \href {http://dx.doi.org/10.1109/ICASSP.2015.7178964}
  {\path{doi:10.1109/ICASSP.2015.7178964}}.
\newline\urlprefix\url{https://doi.org/10.1109/ICASSP.2015.7178964}

\bibitem{busso2008iemocap}
C.~Busso, M.~Bulut, C.~Lee, A.~Kazemzadeh, E.~Mower, S.~Kim, J.~N. Chang,
  S.~Lee, S.~S. Narayanan,
  \href{https://doi.org/10.1007/s10579-008-9076-6}{{IEMOCAP:} interactive
  emotional dyadic motion capture database}, Lang. Resour. Evaluation 42~(4)
  (2008) 335--359.
\newblock \href {http://dx.doi.org/10.1007/s10579-008-9076-6}
  {\path{doi:10.1007/s10579-008-9076-6}}.
\newline\urlprefix\url{https://doi.org/10.1007/s10579-008-9076-6}

\bibitem{nsynth2017}
J.~H. Engel, C.~Resnick, A.~Roberts, S.~Dieleman, M.~Norouzi, D.~Eck,
  K.~Simonyan, \href{http://proceedings.mlr.press/v70/engel17a.html}{Neural
  audio synthesis of musical notes with wavenet autoencoders}, in: D.~Precup,
  Y.~W. Teh (Eds.), Proceedings of the 34th International Conference on Machine
  Learning, {ICML} 2017, Sydney, NSW, Australia, 6-11 August 2017, Vol.~70 of
  Proceedings of Machine Learning Research, {PMLR}, 2017, pp. 1068--1077.
\newline\urlprefix\url{http://proceedings.mlr.press/v70/engel17a.html}

\bibitem{kingma2014adam}
D.~P. Kingma, J.~Ba, \href{http://arxiv.org/abs/1412.6980}{Adam: {A} method for
  stochastic optimization}, in: Y.~Bengio, Y.~LeCun (Eds.), 3rd International
  Conference on Learning Representations, {ICLR} 2015, San Diego, CA, USA, May
  7-9, 2015, Conference Track Proceedings, 2015.
\newline\urlprefix\url{http://arxiv.org/abs/1412.6980}

\bibitem{DBLP:journals/nn/ParisiKPKW19}
G.~I. Parisi, R.~Kemker, J.~L. Part, C.~Kanan, S.~Wermter,
  \href{https://doi.org/10.1016/j.neunet.2019.01.012}{Continual lifelong
  learning with neural networks: {A} review}, Neural Networks 113 (2019)
  54--71.
\newblock \href {http://dx.doi.org/10.1016/j.neunet.2019.01.012}
  {\path{doi:10.1016/j.neunet.2019.01.012}}.
\newline\urlprefix\url{https://doi.org/10.1016/j.neunet.2019.01.012}

\bibitem{DBLP:journals/ijcv/SelvarajuCDVPB20}
R.~R. Selvaraju, M.~Cogswell, A.~Das, R.~Vedantam, D.~Parikh, D.~Batra,
  \href{https://doi.org/10.1007/s11263-019-01228-7}{{Grad-CAM}: Visual
  explanations from deep networks via gradient-based localization}, Int. J.
  Comput. Vis. 128~(2) (2020) 336--359.
\newblock \href {http://dx.doi.org/10.1007/s11263-019-01228-7}
  {\path{doi:10.1007/s11263-019-01228-7}}.
\newline\urlprefix\url{https://doi.org/10.1007/s11263-019-01228-7}

\bibitem{DBLP:conf/icassp/GuizzoWL20}
E.~Guizzo, T.~Weyde, J.~B. Leveson,
  \href{https://doi.org/10.1109/ICASSP40776.2020.9053727}{Multi-time-scale
  convolution for emotion recognition from speech audio signals}, in: 2020
  {IEEE} International Conference on Acoustics, Speech and Signal Processing,
  {ICASSP} 2020, Barcelona, Spain, May 4-8, 2020, {IEEE}, 2020, pp. 6489--6493.
\newblock \href {http://dx.doi.org/10.1109/ICASSP40776.2020.9053727}
  {\path{doi:10.1109/ICASSP40776.2020.9053727}}.
\newline\urlprefix\url{https://doi.org/10.1109/ICASSP40776.2020.9053727}

\bibitem{DBLP:conf/cvpr/WangKFYR19}
H.~Wang, A.~Kembhavi, A.~Farhadi, A.~L. Yuille, M.~Rastegari,
  \href{http://openaccess.thecvf.com/content\_CVPR\_2019/html/Wang\_ELASTIC\_Improving\_CNNs\_With\_Dynamic\_Scaling\_Policies\_CVPR\_2019\_paper.html}{{ELASTIC:}
  improving cnns with dynamic scaling policies}, in: {IEEE} Conference on
  Computer Vision and Pattern Recognition, {CVPR} 2019, Long Beach, CA, USA,
  June 16-20, 2019, Computer Vision Foundation / {IEEE}, 2019, pp. 2258--2267.
\newblock \href {http://dx.doi.org/10.1109/CVPR.2019.00236}
  {\path{doi:10.1109/CVPR.2019.00236}}.
\newline\urlprefix\url{http://openaccess.thecvf.com/content\_CVPR\_2019/html/Wang\_ELASTIC\_Improving\_CNNs\_With\_Dynamic\_Scaling\_Policies\_CVPR\_2019\_paper.html}

\bibitem{DBLP:conf/mlsp/MarchandP16}
U.~Marchand, G.~Peeters, \href{https://doi.org/10.1109/MLSP.2016.7738904}{Scale
  and shift invariant time/frequency representation using auditory statistics:
  Application to rhythm description}, in: F.~A.~N. Palmieri, A.~Uncini, K.~I.
  Diamantaras, J.~Larsen (Eds.), 26th {IEEE} International Workshop on Machine
  Learning for Signal Processing, {MLSP} 2016, Vietri sul Mare, Salerno, Italy,
  September 13-16, 2016, {IEEE}, 2016, pp. 1--6.
\newblock \href {http://dx.doi.org/10.1109/MLSP.2016.7738904}
  {\path{doi:10.1109/MLSP.2016.7738904}}.
\newline\urlprefix\url{https://doi.org/10.1109/MLSP.2016.7738904}

\end{thebibliography}

\end{document}